\documentclass[11pt]{article}
\usepackage[T1]{fontenc}
\usepackage[utf8]{inputenc}
\usepackage{pbox}
\usepackage{booktabs}

\usepackage{tikz}
\usetikzlibrary{shapes}
\usepackage{pgfplots}
\pgfplotsset{width=10cm,compat=1.9}

\usepgfplotslibrary{external}
\usepackage{tikz}
\usetikzlibrary{patterns,decorations.pathreplacing,calc,fadings,calc,shapes.callouts,shapes.arrows,backgrounds,spy,shadows,decorations.markings}
\immediate\write18{mkdir -p tikz}

\usetikzlibrary{intersections}
\usetikzlibrary{positioning}
\usetikzlibrary{decorations.pathreplacing}
\usetikzlibrary{matrix}
\usetikzlibrary{calc}
\usetikzlibrary{arrows,arrows.meta}
\usetikzlibrary{arrows,shapes,trees,backgrounds,shadows}
\usetikzlibrary{decorations.pathreplacing}
\usetikzlibrary{decorations.pathmorphing} 
\usetikzlibrary{backgrounds,snakes}
\usetikzlibrary{decorations.markings}
\usetikzlibrary{positioning}
\usetikzlibrary{plotmarks}
\usetikzlibrary{trees}
\usetikzlibrary{patterns}
\usetikzlibrary{spy}
\usetikzlibrary{calc,matrix}

\newcommand{\n}[2]{n_{#1}^{#2}}
\usepackage{scalerel}

\usepackage{graphicx} 
\usepackage[margin=1in]{geometry}
\usepackage{amsmath}
\usepackage{caption}
\usepackage{hyperref}
\usepackage{subcaption}
\usepackage{xcolor}
\usepackage{amssymb}
\usepackage[numbers]{natbib}
\usepackage{multirow}
\usepackage{soul}

\usepackage{amsmath, amsfonts, dsfont}
\usepackage{amsthm}
\usepackage{hyperref}
\usepackage{graphicx}
\usepackage{placeins}
\usepackage[ruled]{algorithm2e}

\newtheorem{theorem}{Theorem}

\newtheorem{remark}{Remark}

\newcommand{\argmax}{\mathrm{argmax}}

\SetKwInput{KwInputs}{Inputs}
\SetKwInput{KwOutputs}{Outputs}
\SetKwRepeat{Repeat}{repeat}{until}

\newcommand{\Region}{V}
\newcommand{\Size}{N}

\newcommand{\origin}{u}
\newcommand{\destination}{v}
\newcommand{\batteryod}{b_{\origin\destination}}

\newcommand{\range}{B}
\newcommand{\type}{\delta}
\newcommand{\Type}{\Delta}
\renewcommand{\n}[2]{n_{#1}^{#2}}
\newcommand{\region}{v}

\newcommand{\Horizon}{T}
\renewcommand{\time}{t}

\newcommand{\areroutered}[1]{\hat{\mathrm{e}}_{#1}}
\newcommand{\afulfillred}[1]{\hat{\mathrm{f}}_{#1}}
\newcommand{\achargered}[1]{\hat{\mathrm{q}}_{#1}}
\newcommand{\apassred}{\hat{\mathrm{p}}}
\renewcommand{\day}{d}

\newcommand{\totaldays}{D}
\newcommand{\arrrate}[2]{\lambda_{#1}^{#2}}
\newcommand{\arrnum}[2]{X_{#1}^{#2}}

\newcommand{\timecost}[2]{\tau_{#1}^{#2}}

\newcommand{\batterycost}[1]{b_{#1}}
\newcommand{\Lc}{L_c}
\newcommand{\Lp}{L_p}
\newcommand{\chargetime}{J}
\newcommand{\state}[2]{s_{#1}^{#2}}

\newcommand{\timetoarrival}{\eta}
\newcommand{\plugremaintime}{j}
\newcommand{\battery}{b}
\newcommand{\maxtimecost}[1]{\hat{\tau}_{#1}}

\newcommand{\tripactivetime}{\xi}

\newcommand{\cartype}[1]{c_{#1}}
\newcommand{\Cartype}{\mathcal{C}}
\newcommand{\triptype}[1]{o_{#1}}
\newcommand{\Triptype}{\mathcal{O}}
\newcommand{\plugtype}[1]{w_{#1}}
\newcommand{\Plugtype}{\mathcal{W}}
\newcommand{\jointaction}[2]{a^{#2}_{#1}}
\newcommand{\jointtripfulfillaction}[2]{f^{#2}_{#1}}
\newcommand{\jointreroutingaction}[2]{e^{#2}_{#1}}
\newcommand{\jointchargingaction}[2]{q^{#2}_{#1}}
\newcommand{\jointidlingaction}[2]{i^{#2}_{#1}}
\newcommand{\jointpassaction}[2]{p^{#2}_{#1}}

\newcommand{\reward}[2]{r^{#2}_{#1}}
\newcommand{\tripfulfillreward}[2]{r^{#2}_{f #1}}

\newcommand{\reroutingreward}[2]{r^{#2}_{e #1}}
\newcommand{\chargingreward}[2]{r^{#2}_{q #1}}

\newcommand{\lpredtripfulfill}[2]{\bar{f}^{\dagger #2}_{#1}}
\newcommand{\lpredreroute}[2]{\bar{e}^{\dagger #2}_{#1}}
\newcommand{\lpredcharge}[2]{\bar{q}^{\dagger #2}_{#1}}
\newcommand{\lpredslack}[2]{\bar{\mu}^{\dagger #2}_{#1}}
\newcommand{\lpredpass}[2]{\bar{p}^{\dagger #2}_{#1}}

\newcommand{\lptripfulfill}[2]{\bar{f}^{#2}_{#1}}
\newcommand{\lpreroute}[2]{\bar{e}^{#2}_{#1}}

\newcommand{\lpcharge}[2]{\bar{q}^{#2}_{#1}}
\newcommand{\lpslack}[2]{\bar{\mu}^{#2}_{#1}}
\newcommand{\lppass}[2]{\bar{p}^{#2}_{#1}}

\usepackage{xcolor}

\begin{document}




\title{Atomic Proximal Policy Optimization for Electric Robo-Taxi Dispatch and Charger Allocation}


\author{%
    Jim Dai\\
    \footnotesize{Operations Research and Information Engineering, Cornell University, jd694@cornell.edu}\and
    Manxi Wu\\
    \footnotesize{Department of Civil and Environmental Engineering, University of California, Berkeley, manxiwu@berkeley.edu}\and
    Zhanhao Zhang\\
    \footnotesize{Operations Research and Information Engineering, Cornell University, zz564@cornell.edu}
}

\date{}
\maketitle

\begin{abstract}
    Pioneering companies such as Waymo have deployed electric robo-taxi services in several U.S. cities. Operating such fleets requires coordinating ride matching, vehicle repositioning, and battery charging in a stochastic environment while accounting for limited charging infrastructure. Understanding both how to design efficient, scalable control algorithms for large fleets and how charger deployment shapes system performance is therefore critical for robo-taxi operators. 
We model the operations of an electric robo-taxi system as a discrete-time Markov decision process with a long-run average reward objective. To efficiently evaluate system performance under different configurations, we develop a scalable deep reinforcement learning algorithm, Atomic Proximal Policy Optimization (Atomic-PPO), which reduces the action space using atomic action decomposition. The resulting training procedure requires only modest computational resources, making it practical to evaluate a large number of system configurations. We calibrate a simulation environment using real-world NYC for-hire vehicle trip records and benchmark Atomic-PPO against heuristic policies and a fluid-based reward upper bound. The results show that Atomic-PPO produces near-optimal fleet dispatching policies and serves as a reliable tool for system evaluation. We conduct extensive numerical experiments to evaluate how charging infrastructure and fleet size affect system performance. Our findings highlight the importance of ridership-based charger placement, balanced investments in charger quantity and charging speed, and joint planning of charging infrastructure and fleet size.
\color{black}

\end{abstract}



%


\section{Introduction}
Robo-taxi services have been deployed in several U.S. cities, such as San Francisco, Phoenix, Atlanta, and Dallas \citep{robotaxi_news}. 
Efficient operation of an electric robo-taxi fleet is challenging because trip demand is stochastic and varies across time and space, while vehicles must also be routed to charge under limited charging infrastructure.
Poor coordination between fleet operations and charging infrastructure planning can render vehicles unavailable during peak demand periods, leading to reduced service quality and revenue loss. This creates a joint planning challenge for robo-taxi operators: developing efficient fleet control algorithms for large-scale operations while planning charging infrastructure that is well aligned with fleet size and operational demand. 

In this work, we develop a scalable reinforcement learning (RL) algorithm to control large-scale electric robo-taxi fleets for ride matching, vehicle repositioning, and battery charging, and use it to evaluate performance under different configurations of charger deployment and fleet size.
We formulate the fleet operations problem as a Markov decision process (MDP) with discrete state and action spaces. The system state records the spatial distribution of vehicles, their current tasks, active trip requests, and the availability of chargers. Based on the current state, the system dispatches vehicles to fulfill trips, reposition, and charge, earning trip revenues net of repositioning and charging costs. Our objective is to maximize the long run average reward over an infinite time horizon for a given configuration of charging infrastructure and fleet size, which reflects the continuous 24-hour operations of robo-taxi companies.

Computing the fleet dispatching policy is challenging because the state space represents all possible distributions of vehicle statuses and the action space includes all feasible joint vehicle assignments. We develop a scalable deep RL algorithm based on atomic proximal policy optimization (Atomic-PPO), first proposed in \citep{feng2021scalable}. Atomic-PPO decomposes fleet dispatching into sequential assignments of atomic actions -- trip fulfillment, reposition, charge or pass (i.e., no new task) -- to individual vehicles, which we refer to as ``atomic action decomposition". The size of the atomic action space is independent of fleet size. This avoids enumerating combinations of individual vehicle actions, whose number can grow exponentially with fleet size. We integrate atomic action decomposition into Proximal Policy Optimization (PPO) \citep{schulman2017proximal}, which has been effective in policy optimization across various applications \citep{vinyals2019grandmaster, berner2019dota, simm2020reinforcement, zoph2018learning, akkaya2019solving}. 
We approximate the value function and the policy over atomic actions using neural networks. We further reduce the state dimension by clustering vehicle battery levels and grouping trip requests separately by origin region and by destination region.

Furthermore, we derive an upper bound on the optimal long-run average reward using a fluid-based linear program (LP) (see Theorem \ref{thm:fluid-obj-val}). The fluid limit is attained as fleet size approaches infinity, with trip demand and the number of chargers scaling proportionally. This upper bound allows us to assess the trained policy's optimality gap in numerical experiments.

All numerical experiments are conducted in a simulation environment calibrated using the NYC For-Hire Vehicle dataset \citep{nyctlc} for the Manhattan area \footnote{Our code can be found in \url{https://github.com/zhangz73/robo-taxi-dispatch}.}. We consider a fleet of 300 vehicles, comparable to the fleet size deployed by Waymo in a single city \citep{waymo_news}. Trip demand is scaled proportionally to the fleet size. The simulator captures key operational features including nonlinear charging rates and time-varying trip rewards.

To assess the reliability of Atomic-PPO for evaluating system reward, we compare the fleet dispatching policy learned by Atomic-PPO against the fluid-based reward upper bound and two benchmark heuristics. The first heuristic uses randomized rounding to obtain feasible vehicle assignments from the fluid-based linear program, while the second is the power-of-$k$ dispatching policy that processes requests in revenue-descending order and assigns each to the highest-battery vehicle among up to $k$ closest vehicles \citep{varma2023electric}. In the baseline setting with abundant 75 kW DC fast chargers in all regions, Atomic-PPO achieves 91\% of the fluid reward upper bound, whereas the power-of-$k$ policy and the rounded fluid policy achieve 81\% and 69\%, respectively. Atomic-PPO also achieves 89\% of the fluid reward upper bound when vehicle repositioning is disabled throughout the day, indicating that its advantage does not arise simply from allowing vehicle repositioning. Rather, Atomic-PPO selectively fulfills trip requests while accounting for their impact on the future spatial distribution of the fleet and trip demand. Its strong performance, both with and without vehicle repositioning, supports using Atomic-PPO to evaluate system reward under different charging infrastructure and fleet configurations. Training a policy for 300 vehicles takes four hours on 30 CPU cores.

We further study how charger deployment and fleet size affect system reward, yielding several managerial insights. First, charger placement matters: 15 chargers (i.e. 5\% of fleet size) concentrated in Midtown Manhattan achieve approximately the same average reward as 30 chargers (i.e. 10\% of fleet size) uniformly allocated across all regions.
Second, both the number of chargers and the charging speed affect system reward. When either charger quantity or charging speed is low, increasing the other improves reward by allowing vehicles to return to service more quickly. As either becomes sufficiently high, further increases yield smaller improvements. These results suggest that charger quantity and charging speed can partly substitute for one another and should be considered together. 
Finally, fleet size and charging infrastructure should be planned together: smaller fleets reach diminishing returns from additional charging capacity at smaller charger quantities or lower outlet rates, whereas larger fleets benefit from further increases in charger quantity or outlet rate.


We summarize our \textbf{main contributions} below.
\begin{enumerate}
    \item \textbf{Scalable Algorithm}: We develop an efficient algorithm, Atomic-PPO, that jointly optimizes trip matching, vehicle repositioning, and charging, and is able to compute near-optimal fleet dispatching policies using moderate computational resources.
    \item \textbf{Fluid-Based Upper Bound}: We derive a fluid-based upper bound on the optimal long-run average reward, which can be used to assess the optimality gap of any fleet dispatching policy.
    \item \textbf{Managerial Insights}: Leveraging Atomic-PPO, we conduct extensive numerical experiments to evaluate system performance under different configurations of charger deployment and fleet size. Our experiments provide several managerial insights on how charger placement, charger quantity, charging speed, and fleet size affect system reward, offering guidance on efficient planning of charging infrastructure and fleet operations for electric robo-taxi systems.
\end{enumerate}

\color{black}

\paragraph{Related Literature.} Optimal fleet dispatch has been extensively studied in ride-hailing systems, mostly for gasoline vehicles without charging constraints, with several recent works on electric vehicles (EVs) \citep{varma2023electric, roberti2016electric, luke2021joint, boewing2020vehicle, zalesak2021real, shi2019operating, kullman2022, dean2022synergies, yu2023coordinating, singh2024dispatching, dong2022dynamic, su2024branch, ahadi2023cooperative, liu2022smart, gao2024stochastic, li2024bm, zhu2023dynamic, al2020approximate, iacobucci2019optimization, turan2020dynamic, song2022sumo}. These methods fall into three categories: (1) deterministic optimization methods; (2) queueing-based analysis; and (3) deep reinforcement learning. 

\medskip 
\noindent(1) Deterministic optimization. With all future trip arrivals known, optimal fleet dispatch can be formulated as a linear program \citep{luke2021joint} or a mixed-integer program (MIP) \citep{zalesak2021real, boewing2020vehicle, dean2022synergies, tuncel2023integrated, yu2023coordinating, su2024branch, li2024bm, iacobucci2019optimization}. These formulations do not directly account for uncertain trip demand. 

\medskip 
\noindent(2) Queueing-Based Analysis. Some studies model ride-hailing as a closed queueing system \citep{banerjee2018, braverman2019empty, zhang2018modeling, waserhole2012vehicle, afeche2023}, where drivers remain in the system indefinitely. Others use an open queueing system \citep{varma2023electric, dong2022dynamic, wang2022, ozkan2020joint, ozkan2020dynamic, xu2021generalized}, where drivers dynamically enter and exit the system.

All of these studies except for \citet{varma2023electric} focus on gasoline vehicle dispatching and do not consider charging. Moreover, most assume time-homogeneity, with exceptions such as \citep{ozkan2020joint, ozkan2020dynamic, afeche2023, xu2021generalized}, which incorporate time-varying dynamics. These studies develop fleet control policies that are optimal under specific conditions or in the fluid limit. Our construction of a reward upper bound builds on the fluid-based analytical methods developed in this literature. However, optimality in the fluid limit does not ensure a small optimality gap for moderate fleets. In particular, we find that the power-of-k policy proposed in \cite{varma2023electric} achieves a lower reward than our policy in our experiments.

\medskip 
\noindent(3) Deep Reinforcement Learning. The third category of research uses deep reinforcement learning (RL) to train fleet dispatching policies in complex environments that incorporate both spatial and temporal demand uncertainty. For a comprehensive review of this literature, see the recent survey by \cite{qin2022reinforcement}. Some studies use a semi-Markov decision process (semi-MDP), with decisions triggered by trip arrivals or vehicle task completions \citep{kullman2022, singh2024dispatching, gao2024stochastic}. In contrast, the majority of studies in this category, including ours, model the ride-hailing system as a discrete-time MDP, where dispatch decisions are made for the entire fleet at discrete time steps.



Several studies adopt hierarchical frameworks to reduce the {complexity of fleet dispatching}. One line of work partitions decisions across regions (or grids), where the upper level determines the fraction of vehicles assigned to each action for all vehicles within a region, while the lower level assigns those actions to individual vehicles based on their status (e.g., battery level) \citep{zhu2023dynamic, xie2023two, si2024vehicle, jin2019coride}. Another approach decomposes the problem into separate subproblems such as ride matching, vehicle repositioning, and charging. For example, \cite{gammelli2022graph, schmidt2024learning, tresca2025robo, singhal2024real} first match available vehicles to trip requests to maximize immediate rewards, and then assign the remaining vehicles to repositioning or charging while accounting for future value. Although some studies attempt to address these subproblems jointly, many focus on only one or two components rather than providing an integrated framework. A few exceptions include \cite{al2020approximate, turan2020dynamic}. Our model also jointly controls ride matching, repositioning, and charging.


Prior work has adopted the $T$-step look-ahead approach \citep{wei2023reinforcement}, which computes the optimal fleet dispatching actions for the first $T$ steps of value iteration by solving a linear program, and the warm-start approach \citep{yuan2022reinforcement}, which initializes the fleet dispatching policy through imitation learning from mixed-integer programs. Both remain computationally challenging because they repeatedly solve large-scale linear or mixed-integer programs. Some studies reduce the complexity by restricting vehicles within the same region to take identical actions \citep{qin2021optimizing, liu2022deep}. Others model fleet actions as continuous variables or round them to discrete decisions \citep{mao2020dispatch, zhou2023robust, turan2020dynamic, gammelli2022graph, skordilis2021modular, schmidt2024learning, xie2023two, si2024vehicle}. Some studies instead train a deep Q-network to compute scores for each region based on the current state and assign idle vehicles to reposition according to these scores in a round-robin manner \citep{liu2022deep}. These approximations may introduce errors, particularly when ride matching, repositioning, and charging are considered jointly. The studies by \cite{gammelli2022graph, schmidt2024learning, tresca2025robo, singhal2024real} instead compute the target number of idle vehicles in each region using RL-based methods, and then assign repositioning actions to satisfy these targets while minimizing repositioning costs. {This avoids directly computing flows across all origin–destination pairs, although the number of possible regional vehicle-count targets still grows with fleet size.}

In a decentralized reinforcement learning framework, vehicles, vehicle–action pairs, or groups of vehicles within regions are treated as agents that make decentralized decisions. In \cite{si2024vehicle, wang2018deep, xi2022hmdrl, chen2024rebalance, xi2021ddrl, wen2017rebalancing, singh2021distributed, liu2022smart}, each agent acts independently based on its own Q-function, which evaluates the expected cumulative reward associated with its individual actions. This approach avoids enumerating the exponentially large joint action space but does not capture coordination across agents. Two approaches address this issue: (i) learning a Q-function that incorporates joint-action effects by using a mean-field approximation for the actions of all other agents except the current one \citep{zhu2023dynamic, xie2023two, li2019efficient, ahadi2023cooperative}; and (ii) computing a score for each vehicle–action pair (e.g., using an individual Q-function) and assigning actions to agents through a bipartite matching that maximizes the sum of these scores \citep{xu2018large, tang2019deep, qin2020ride, enders2023hybrid}. However, the resulting policy remains theoretically suboptimal because neither approach fully accounts for the collective impact of individual agent actions on state transitions and overall system reward. Our Atomic-PPO algorithm addresses this issue by introducing a coordinated sequential assignment of atomic actions that incorporates state transitions after each step. This reduces the action space while capturing each vehicle’s impact on the system state and learning the accumulated rewards from fleet dispatch decisions at each decision epoch. The concurrent paper \citep{dai2025optimal} establishes the theoretical foundation showing that the atomic action decomposition preserves optimality.


Our idea of atomic action decomposition is related to cooperative multi-agent reinforcement learning (MARL), where multiple agents act based on a shared global state to maximize a common reward, with both the reward and state transitions determined by their joint actions \citep{kuba2021trust, bertsekas2019multiagent, gu2021multi, kuba2022heterogeneous, wang2023order, kuba2021settling, wen2022multi, fu2022revisiting, ye2022towards, wen2024reinforcing}. The joint action space can grow exponentially with the number of agents, making policy training challenging. A common strategy, referred to as {\em auto-regressive policies}, is to factor the joint policy into a sequence of lower dimensional policies, where each agent’s action is generated conditioned on the actions of preceding agents \citep{wen2022multi, fu2022revisiting, ye2022towards, wen2024reinforcing}.

Our atomic policy framework exploits the structure of fleet dispatching to improve scalability. Instead of conditioning each vehicle’s action on the entire history of preceding actions, it conditions on the updated system state. Each atomic action induces explicit changes in vehicle status, allowing this state update to capture its effects without retaining the action history. Consequently, the input dimension of the atomic policy remains constant and independent of fleet size, while the input dimension of auto-regressive policies {that explicitly append preceding actions} scales with the number of agents.

Finally, our Atomic-PPO algorithm builds on \citep{feng2021scalable}, which addressed ride matching and vehicle repositioning for conventional vehicle fleets. We extend that framework to vehicle charging and introduce a new state reduction scheme, enabling training for realistic problem sizes using moderate computational resources. Furthermore, we provide a theoretical upper bound on the optimal long-run average reward, which serves as a benchmark for evaluating fleet dispatching policies. Atomic-PPO achieves 91\% of this bound in the baseline experiments, supporting its use for evaluating system reward. Its computational efficiency enables extensive experiments on charger deployment and fleet size to obtain managerial insights.
\color{black}

\section{Model} \label{sec:model}
\subsection{The ride-hailing system}\label{subsec:ride-hail}
We consider a transportation network with a set $\Region$ of service regions. In this network, a fleet of electric vehicles with size $\Size$ are operated by a central planner to serve passengers, who make trip requests from their origins $\origin \in \Region$ to their destinations $\destination \in \Region$. For each pair of $(\origin, \destination) \in \Region \times \Region$, we assume that the battery consumption for traveling from $\origin$ to $\destination$ is a constant $\batteryod \in \mathbb{R}_{\geq 0}$. The electric vehicles are each equipped with a battery of size $\range$, i.e. a fully charged vehicle can take trips with total battery consumption up to $\range$. A set of chargers with different charging rates $\type \in \Type$ are installed in the network, where each rate $\type$ charger can replenish $\delta$ amount of battery to the connected vehicle in one unit of time. We denote the number of rate $\type \in \Type$ chargers in each region $\region \in \Region$ as $\n{\region}{\type} \in \mathbb{N}_{\geq 0}$.



We model the operations of a ride-hailing system as a discrete-time, average reward Markov Decision Process (MDP) with infinite time horizon. In particular, we model the system as an infinite repetition of a single-day operations, where each day $\day =1, 2, \dots$ is evenly partitioned into $\Horizon$ number of time steps $\time =1, \dots, \Horizon$. The system makes a dispatching decision for the entire fleet at every time step of a day, which we refer to as a decision epoch. In each decision epoch $(\time, \day)$, the number of trip requests between each $\origin$-$\destination$ pair, denoted as $\arrnum{\origin\destination}{\time, \day}$, follows a Poisson distribution with mean $\arrrate{\origin\destination}{\time}$. The duration of trips from region $\origin$ to region $\destination$ at time $t$ is a constant $\timecost{\origin\destination}{\time}$ that is a multiple of the length of a time step. Both $\arrrate{\origin\destination}{\time}$ and $\timecost{\origin\destination}{\time}$ can vary across time steps in a single day, but remain unchanged across days.

When the central planner receives the trip requests, they assign vehicles to serve (all or a part of) the requests. In particular, a vehicle must be assigned to the passenger within the connection patience time $\Lc \geq 0$, and a passenger will wait at most time $\Lp \geq 0$ (defined as the pickup-patience time) for an assigned vehicle to arrive at their origin. Otherwise, the passenger will leave the system. Both $\Lp$ and $\Lc$ are multiples of the discrete time steps.

The central planner keeps track of the status of each vehicle by the region $\region \in \Region$ it is currently located or heading to, remaining time to reach $\timetoarrival = 0, \dots, \maxtimecost{\region}$, and remaining battery level $\battery =0, \dots, \range$ when reaching $\region$. Here, $\maxtimecost{\region} : = \max_{\origin \in \Region, \time \in [\Horizon]} \timecost{\origin \region}{\time}$ is the maximum duration of any trip with destination $\region$. We assume that the minimum time cost of all trips $\min_{\origin,\destination \in \Region, \time \in [\Horizon]} \timecost{\origin\destination}{\time}$ is larger than $\Lp$ so that no vehicle can be assigned to serve more than one trip request in a single decision epoch.

A vehicle is associated with status $\cartype{}= (\region, \timetoarrival, \battery)$ if (1) it is routed to destination $\region$, with remaining time $\timetoarrival$, and battery level $\battery$ at the arrival; or (2) it is being charged at region $\region$, with remaining charging time $\timetoarrival$, and battery level $\battery$ after the charging period completes. Here, if $\timetoarrival=0$, then the vehicle is idling at region $\region$. Additionally, if $\timetoarrival > 0$, then the vehicle could either be taking a trip with a passenger whose destination is $\region$, being repositioned to $\region$, or being charged in $\region$. Vehicle repositioning refers to sending an idle vehicle to another region without carrying a passenger, where it may subsequently serve trip requests, charge, or idle. We note that the maximum remaining time $\eta$ for trip fulfilling vehicles cannot exceed a finite number $\maxtimecost{\region} + \Lp$ since a vehicle is only eligible to serve a trip if the vehicle can arrive at the origin of that trip within $\Lp$ time and the maximum trip duration is $\maxtimecost{\region}$. Additionally, a vehicle with status $(\region, 0, \battery)$ can be charged at region $\region$ with rate $\type$ if such a charger is available in $\region$.\footnote{We can extend our model to account for non-linear charging rates, see our numerical experiment in Sec. \ref{sec:numerical}.} Charging takes $\chargetime$ time steps. If the vehicle is charged to full within $\chargetime$ time steps, then it will remain connected to the charger and idle for the remaining time of the charging period. We assume that $\chargetime > \Lp$ so that vehicles assigned to charge will not be assigned to serve trip requests in the same decision epoch. 
Let $\Cartype := \left\{(\region, \timetoarrival, \battery) \right\}_{\region \in \Region, \timetoarrival = 0, \dots, \maxtimecost, \battery = 0, \dots, \range}$ denote the set of all vehicle status. The set $\Cartype$ is finite. 
 
\color{black}
A trip order is associated with status $\triptype{}= (\origin, \destination, \tripactivetime)$ if it originates from $\origin$, heads to $\destination$, and has been waiting for vehicle assignment in the system for $\tripactivetime$ time steps. A trip with origin $\origin$ and destination $\destination$ can be served by a vehicle of status $(\origin, \timetoarrival, \battery)$ if (1) the vehicle can reach $\origin$ within the passenger's pickup-patience time $\Lp$ (i.e. $\eta \leq \Lp$), and (2) the remaining battery of the vehicle when reaching $\origin$ is sufficient to complete the trip to $\destination$ (i.e. $b \geq \batteryod$). We note that a vehicle may be assigned to pick up a new passenger before completing its current trip towards $\origin$ as long as it can arrive within $\Lp$ time steps. We use $\Triptype := \left\{(\origin, \destination, \tripactivetime) \right\}_{\origin, \destination\in \Region,\ \tripactivetime = 0, \dots, \Lc}$ to denote the set of all trip status. 

A charger is associated with status $\plugtype{}= (\region, \type, \plugremaintime)$ if it is located at region $\region$ with a rate of $\type$ and is $\plugremaintime$ time steps away from being available. Specifically, if $\plugremaintime = 0$, the charger is available immediately. If $\plugremaintime > 0$, the charger is currently in use and will take an additional $\plugremaintime$ time steps to complete the charging period. We use $\Plugtype := \{(\region, \type, \plugremaintime)\}_{\region \in \Region, \type \in \Type, \plugremaintime = 0, \dots, \chargetime - 1}$ to denote the set of all charger status. The main notation is summarized in Table \ref{tab:notations} in Appendix \ref{sec:appendix-notations}.

\subsection{Markov decision process} \label{subsec:MDP}
We next describe the \emph{state, action, policy} and \emph{reward} of the Markov decision process (MDP).

\paragraph{State.} We denote the state space of the MDP as $\mathcal{S}$ with generic element $\state{}{}$. The state vector $s^{t, d}= \left(t, \left(\state{\cartype{}}{\time,\day} \right)_{\cartype{} \in \Cartype}, \left(\state{\triptype{}}{\time,\day} \right)_{\triptype{} \in \Triptype}, \left(\state{\plugtype{}}{\time,\day} \right)_{\plugtype{} \in \Plugtype} \right) \in \mathcal{S}$ records the fleet state $\left(\state{\cartype{}}{\time,\day} \right)_{\cartype{} \in \Cartype}$, the trip order state $\left(\state{\triptype{}}{\time,\day} \right)_{\triptype{} \in \Triptype}$, and the charger state $\left(\state{\plugtype{}}{\time,\day} \right)_{\plugtype{} \in \Plugtype}$ for each time $\time$ of day $\day$, where $\state{\cartype{}}{\time,\day}$ is the number of vehicles of status $\cartype{}$, $\state{\triptype{}}{\time,\day}$ is the number of trip orders of status $\triptype{}$, and $\state{\plugtype{}}{\time, \day}$ is the number of chargers of status $\plugtype{}$ at $(t, d)$. For all $(t, d)$, the sum of vehicles of all statuses equals to the fleet size $\Size$. That is, 
\begin{equation} \label{eq:setup-car-total-flow}
    \sum_{\cartype{} \in \Cartype} \state{\cartype{}}{\time, \day} = \Size,\quad \forall \time \in [\Horizon],\ \day = 1, 2, \dots.
\end{equation}
Additionally, the sum of chargers of all remaining charging times for each region $\region$ and rate $\type$ equals to the quantity of the corresponding charging facility $\n{\region}{\type}$. That is, for all $\region \in \Region$ and all $\type \in \Type$,
\begin{equation} \label{eq:setup-charger-total-num}
    \sum_{j=0}^{J} \state{(\region, \type, \plugremaintime)}{\time, \day} = \n{\region}{\type},\quad \forall \time \in [\Horizon],\ \day = 1, 2, \dots.
\end{equation}
Moreover, we note that while the number of new trip arrivals $\arrnum{\origin\destination}{\time, \day}$ can be unbounded, the maximum number of trip orders that can be served before abandonment at any time step is at most $\Size(\Lc + 1)$. This is because the fleet can serve up to $\Size$ trips per time step, and each trip order remains active for at most $\Lc$ steps before being abandoned. As a result, without loss of generality, we can upper bound $\state{\triptype{}}{\time,\day}$ for each status $\triptype{}$ by $\Size(\Lc + 1)$, with any additional trip requests beyond this limit rejected by the system. Hence, our state space $\mathcal{S}$ is finite.

\paragraph{Action.} We denote the action space of the MDP as $\mathcal{A}$ with generic element $\jointaction{}{}$. An action is a flow vector of the fleet that consists of the number of vehicles of each status assigned to take trips, reposition, charge, and pass. 
We denote an action vector at time $\time$ of day $\day$ as $\jointaction{}{\time,\day} := \left(\jointtripfulfillaction{\cartype{}}{\time,\day}, \jointreroutingaction{\cartype{}}{\time,\day}, \jointchargingaction{\cartype{}}{\time,\day}, \jointpassaction{\cartype{}}{\time,\day}\right)_{\cartype{} \in \Cartype}$, where: 

\begin{itemize}
    \item[-] \emph{$\jointtripfulfillaction{\cartype{}}{\time,\day} := \left(\jointtripfulfillaction{\cartype{}, \triptype{}}{\time,\day} \in \mathbb{N} \right)_{\triptype{} \in \Triptype}$} determines the number of vehicles of each status $\cartype{}$ assigned to fulfill each trip order status $\triptype{}\in \Triptype$ at time $\time$ of day $\day$. In particular, vehicles are eligible to take trip requests if their current location or destination matches the trip's origin region, they are within $\Lp$ time steps from completing the current task, and they have sufficient battery to complete the trip, i.e.  
    \begin{align} \label{eq:setup-passenger-carrying-flow}
        \jointtripfulfillaction{\cartype{}, \triptype{}}{\time,\day} \left\{
        \begin{array}{ll}
            \geq 0, & \quad \text{$\forall \cartype{} = (\origin, \timetoarrival, \battery)$ and $\triptype{} = (\origin, \destination, \tripactivetime)$}\\
            &\text{ such that $\timetoarrival \leq L_p$ and 
            $\battery \geq \batterycost{\origin\destination} + \batterycost{\destination \destination}$}, \\
            = 0, & \quad \text{otherwise.}
        \end{array}
        \right.
    \end{align}
    Additionally, we require that the total number of vehicles that fulfill the trip orders of status $\triptype{}$ cannot exceed $\state{\triptype{}}{\time, \day}$, i.e. 
    \begin{equation} \label{eq:setup-trip-fulfill-cap}
        \sum_{\cartype{} = (\origin, \timetoarrival, \battery) \in \Cartype} \jointtripfulfillaction{\cartype{}, \triptype{}}{\time,\day} \leq \state{\triptype{}}{\time, \day},\quad \forall \triptype{} = (\origin, \destination, \tripactivetime) \in \Triptype.
    \end{equation}
    \item[-] \emph{$\jointreroutingaction{\cartype{}}{\time,\day} := \left(\jointreroutingaction{\cartype{}, \destination}{\time,\day} \in \mathbb{N} \right)_{\destination \in \Region}$} represents the number of vehicles of each status $\cartype{}$ assigned to reposition to $\destination$ at time $\time$ of day $\day$. In particular, vehicles are eligible to reposition to a different region if they have already completed their current tasks and they have sufficient battery to complete the trip. 
    \begin{align} \label{eq:setup-rerouting-flow}
        \jointreroutingaction{\cartype{}, \destination}{\time,\day} \left\{
        \begin{array}{ll}
            \geq 0, & \quad \text{$\forall \cartype{} = (\origin, \timetoarrival, \battery)$ and $\destination \neq \origin$}\\
            &\text{ such that $\timetoarrival = 0$ and 
            $\battery \geq \batterycost{\origin \destination} + \batterycost{\destination \destination}$}, \\
            = 0, & \quad \text{otherwise.}
        \end{array}
        \right.
    \end{align}
    \item[-] \emph{$\jointchargingaction{\cartype{}}{\time,\day} := \left(\jointchargingaction{\cartype{}, \type}{\time,\day} \in \mathbb{N} \right)_{\type \in \Type}$} represents the number of vehicles of status $\cartype{}$ assigned to charge with rate $\type$ at time $\time$ of day $\day$. In particular, vehicles are eligible to charge if they are idling in the region where the charger is located: for each $\cartype{} = (\origin, \timetoarrival, \battery)$ and $\type \in \Type$, $\jointchargingaction{\cartype{}, \type}{\time,\day} \geq 0$ if $\timetoarrival = 0$, and $\jointchargingaction{\cartype{}, \type}{\time,\day} = 0$ otherwise.
    Additionally, for each region $\region$ and charging rate $\type$, we require that the total number of vehicles of all statuses assigned to charge at region $\region$ with rate $\type$ cannot exceed the total number of available chargers:
       {\begin{equation} \label{eq:setup-charging-cap}
        \sum_{\cartype{} = (\region, \timetoarrival, \battery) \in \Cartype} \jointchargingaction{\cartype{}, \type}{\time,\day} \leq \state{(\region, \type, 0)}{\time, \day},\quad \forall \region \in \Region,\  \type \in \Type.
    \end{equation}}
    \item[-] \emph{$\jointpassaction{\cartype{}}{\time,\day} \in \mathbb{N}$} represents the number of vehicles of status $\cartype{}$ not assigned with any new action at time $\time$ of day $\day$. All vehicles are eligible for the pass action. 
\end{itemize}

For any $\time$ and $\day$, the vector $\jointaction{}{\time, \day}$ needs to satisfy the following flow conservation constraint: All vehicles of each status should be assigned to one of the trip-fulfilling, repositioning, charging, or passing actions. That is,
\begin{align} \label{eq:setup-cartype-conservation}
    &\sum_{\triptype{} \in \Triptype} \jointtripfulfillaction{\cartype{}, \triptype{}}{\time,\day} + \sum_{\destination \in \Region} \jointreroutingaction{\cartype{}, \destination}{\time,\day} + \sum_{\type \in \Type} \jointchargingaction{\cartype{}, \type}{\time,\day} + \jointpassaction{\cartype{}}{\time,\day} = \state{\cartype{}}{\time,\day}, \notag\\
    &\qquad \forall \cartype{} \in \Cartype,\ \time \in [\Horizon],\ \day = 1, 2, \dots.
\end{align}
From the above description, we note that the feasibility of an action depends on the state. We denote the set of actions that are feasible for state $s$ as $\mathcal{A}_s$. 

\paragraph{Policy.} The policy $\pi: \mathcal{S} \rightarrow \Delta(\mathcal{A}_s)$ is a stochastic policy that maps the state vector to an action, where $\pi(\jointaction{}{} \vert \state{}{})$ is the probability of choosing action $\jointaction{}{}$ given state $\state{}{}$ under policy $\pi$. We note that the notation $\pi(\cdot)$ does not explicitly depend on $t$ because $t$ is already included as a part of the state vector $s$.  

\paragraph{State Transitions.} Given the current state and action, vehicle destinations, remaining times to arrival, and battery levels are updated according to the assigned trip-fulfilling, repositioning, or charging actions. Vehicles assigned to charge first travel to the charger, consuming time and battery, and their battery gains are capped at full charge. Vehicles assigned to pass remain idle if they have no ongoing task; otherwise, their remaining task time decreases by one. Trip orders assigned to vehicles leave the queue; unassigned orders age by one time step and are abandoned once their active time exceeds $\Lc$. New trip orders enter with active time zero, subject to the arrival cap described above. Chargers become available when their charging periods end, while newly assigned charging vehicles occupy available chargers. The complete transition equations are provided in Appendix \ref{apx:state-transitions}.

\paragraph{Reward.} The reward of fulfilling a trip request from $\origin$ to $\destination$ at time $t$ is $\tripfulfillreward{,\origin\destination}{\time} \in \mathbb{R}_{\geq 0}$. We remark that if a trip request has been active in the system for some time, then the reward is determined by the time at which the trip request is picked up. The reward (cost) of vehicle repositioning between $\origin, \destination$ is $\reroutingreward{,\origin\destination}{\time} \in \mathbb{R}_{\leq 0}$. The reward (cost) for a vehicle to charge at time $\time$ is $\chargingreward{,\type}{\time} \in \mathbb{R}_{\leq 0}$. Idling and passing actions incur no reward or cost. As a result, given the action $\jointaction{}{\time, \day}$, the total reward at time $\time$ of day $\day$ is given by 
 \begin{align} \label{eq:setup-reward}
     \reward{}{\time}(\jointaction{}{\time,\day}) =& 
    \sum_{\cartype{} \in \Cartype} \sum_{(\origin, \destination) \in \Region \times \Region}\sum_{\tripactivetime = 0}^{\Lc} \tripfulfillreward{,\origin\destination}{\time} \jointtripfulfillaction{\cartype{}, (\origin, \destination, \xi)}{\time,\day} \notag \\ 
    +& \sum_{\cartype{} \in \Cartype} \sum_{(\origin, \destination) \in \Region \times \Region}
    \reroutingreward{,\origin\destination}{\time} \jointreroutingaction{\cartype{}, \destination}{\time,\day} + \sum_{\cartype{} \in \Cartype}\sum_{\type \in \Type} \chargingreward{,\type}{\time} \jointchargingaction{\cartype{},\type}{\time,\day}.
 \end{align}

The long-run average daily reward of a policy $\pi$ given some initial state $\state{}{}$ is as follows:
\begin{equation} \label{eq:long-run-avg-reward}
    R(\pi \vert \state{}{}) := \lim_{\totaldays \rightarrow \infty} \frac{1}{\totaldays} \mathbb{E}_{\pi}\left[\sum_{\day = 1}^{\totaldays} \sum_{\time = 1}^{\Horizon} \reward{}{\time}(\jointaction{}{\time,\day}) \Biggm\lvert \state{}{} \right].
\end{equation}

Since the state is finite and the state transition and policy are homogeneous across days, the limit defined above exists (see Proposition 8.1.1 on page 333 of \cite{PutermanMDP}). Our goal is to find the optimal fleet control policy $\pi^*$ that maximizes the long-run average daily rewards \footnote{{The long run average reward objective provides a natural formulation for robo-taxi systems operating continuously over a 24 hour horizon. However, we note that this formulation is not essential, as it can be approximated by a discounted reward objective with a discount factor close to 1.}} given {\em any} initial state $\state{}{} \in \mathcal{S}$, i.e. $R^*(\state{}{}) = R(\pi^* \vert \state{}{}) = \max_{\pi} R(\pi \vert \state{}{})$.

\subsection{Model limitations}
First, to ensure tractability and scalability, the service area is partitioned into regions. This spatial aggregation abstracts away fine-grained geographic details, such as traffic signals, one-way streets, and turning constraints that can be important for exact vehicle-to-request matching.
Our work focuses on higher-level, region-to-region fleet dispatching decisions, including vehicle routing to chargers under limited charging capacity and vehicle repositioning to balance spatial demand. Numerical experiments in Appendix \ref{sec:appendix-robustness} indicate that the algorithm continues to perform well with more regions and modest increases in computational resources; Section \ref{sec:deep-rl} discusses its computational complexity.

Second, for notational simplicity in expressing the state transition equations, we model trip durations, charging times, and customer patience times as deterministic, without explicitly modeling heterogeneous patience across customers. The framework can incorporate stochastic versions of these variables or heterogeneous patience. For example, stochastic abandonment can be modeled by allowing each waiting trip request to exit with a certain probability at each time step. Our framework for computing optimal fleet dispatching policies can still be applied in such settings. In the experiments in Appendix \ref{sec:appendix-robustness}, Atomic-PPO achieves comparable average daily rewards under deterministic and stochastic trip durations.

Third, for modeling simplicity, we assume that system parameters such as trip durations and passenger arrival rates are constant across days, omitting day of the week variation and longer term seasonal demand shifts. Day of the week effects can be incorporated by augmenting the state vector with an indicator. With respect to
longer term distribution shifts, our experiments in Appendix \ref{sec:appendix-robustness} show that the trained policy can tolerate moderate month to month changes in demand.

Finally, vehicle repositioning actions are modeled as uninterrupted until completion. In practice, vehicles may dynamically adjust their routes in response to new information or incoming requests, and allowing such interruptions could improve trip fulfillment and system performance. This extension can be incorporated by tracking intermediate vehicle locations in the state, but is left for future work to preserve notational simplicity.

\color{black}

\section{Fleet Control Policy Reduction With Atomic Actions} \label{sec:atomic-action}
One challenge of computing the optimal control policy lies in the size of the action space $\vert\mathcal{A}\vert$. The focus of this section is to address this challenge by introducing a policy reduction scheme, which decomposes fleet dispatching to sequential assignment of tasks to individual vehicles, where the task for each individual vehicle is referred as an ``atomic action". We use the name ``atomic action policy" because each atomic action only changes the status of a single vehicle. In particular, for any vehicle of a status $\cartype{} \in \Cartype$, an atomic action can be any one of the followings: 
\begin{itemize}
    \item[-] $\afulfillred{\triptype{}}$ represents fulfilling a trip of status $\triptype{} \in \Triptype$.
    \item[-] $\areroutered{\destination}$ represents repositioning to destination $\destination \in \Region$.
    \item[-] $\achargered{\type}$ represents charging with rate $\type \in \Type$ at its current region.
    \item[-] $\apassred$ represents continuing with its previously assigned actions. 
\end{itemize}
We use $\hat{\mathcal{A}}$ to denote the atomic action space that includes all of the above atomic actions, i.e. $\hat{a} \in \hat{\mathcal{A}} := \left\{\left(\afulfillred{\triptype{}}\right)_{\triptype{} \in \Triptype}, \left(\areroutered{\destination}\right)_{\destination \in \Region}, \left(\achargered{\type}\right)_{\type \in \Type}, \apassred \right\}$. The atomic action significantly reduces the {policy output dimension} since $\hat{\mathcal{A}}$ does not scale with the {fleet size.} 

We now present the procedure of atomic action assignment. In each decision epoch $(t, d)$, vehicles are arbitrarily indexed from $1$ to $\Size$, and are sequentially selected. For a selected vehicle $n$, the atomic policy $\hat{\pi}: \mathcal{S} \times \Cartype \rightarrow \Delta(\hat{\mathcal{A}})$ maps from the tuple of system state $s^{t,d}_n$ before the $n$-th assignment and the selected vehicle's status $\cartype{n}$ to a distribution of atomic actions. The system state $s^{t,d}_n$ updates after each individual vehicle assignment, starting with $s^{t,d}_1 = s^{t,d}$. After the final vehicle is assigned and the new trip arrivals at time $t+1$ are realized, the state $s^{t,d}_{\Size}$ transitions to $s^{t+1,d}$.

The total reward for each decision epoch $(t, d)$ is the sum of all rewards generated from each atomic action assignment in $(t, d)$, where the reward generated by the atomic action $\hat{a}^{t,d} \in \hat{\mathcal{A}}$ on a vehicle of status $\cartype{}^{t,d} := (\origin, \timetoarrival, \battery) \in \Cartype$ is given by
\begin{align*}
    r^t(\cartype{}^{t,d}, \hat{a}^{t,d}) =& \sum_{\triptype{} \in \Triptype} r^t_{f, \triptype{}}\mathds{1}\left\{\hat{a}^{t,d} = \afulfillred{\triptype{}}\right\}\\
    +& \sum_{(\origin,\destination) \in \Region \times \Region} r^t_{e, \origin\destination}\mathds{1}\left\{\hat{a}^{t,d} = \areroutered{\destination}\right\}\\ 
    +& \sum_{\delta \in \Delta} r^t_{q, \delta}\mathds{1}\left\{\hat{a}^{t,d} = \achargered{\delta}\right\}.
\end{align*}
The long-run average reward given the atomic action policy $\hat{\pi}$ and the initial state $s \in \mathcal{S}$ is as follows: 
\begin{align*}
    &R(\hat{\pi} \vert s) \\
    =& \lim_{\totaldays \rightarrow \infty} \frac{1}{\totaldays} \mathbb{E}_{\hat{\pi}}\left[ \sum_{\day = 1}^{\totaldays}\sum_{t=1}^T \sum_{n=1}^{N} r^{t}(\cartype{n}^{t,d}, \hat{a}^{\time, \day}_{n}) \Bigg\vert s \right],\\
    &\quad\quad \forall s \in \mathcal{S},
\end{align*}
where $\hat{a}_n^{t, d}$ is the atomic action generated by the atomic action policy and $\cartype{n}^{t,d}$ is the status of vehicle to be assigned in the $n$-th atomic step in decision epoch $(t, d)$. Given any initial state $s \in \mathcal{S}$, the optimal atomic action policy is given by $\hat{\pi}^{*} = \argmax_{\hat{\pi}} R(\hat{\pi} \vert s)$. 

Our atomic action policy can be viewed as a reduction of the original fleet dispatching policy in that any realized sequence of atomic actions corresponds to a feasible fleet dispatching action with the same reward of the decision epoch. This reduction makes the training of atomic action policy scalable because the output dimension of atomic action policy $\hat{\pi}^t$ equals to $\vert \hat{\mathcal{A}}\vert$, which is a constant regardless of the fleet size. 

\begin{remark}
    Our atomic policy {avoids the growing input dimension of auto-regressive policies that explicitly append preceding actions.} Under the atomic policy, the input for the $n$-th vehicle depends only on the current system state $s_n^{t,d}$ and the selected vehicle status $c_n^{t,d}$. In contrast, {such} auto-regressive policies require augmenting the input with the atomic actions of all preceding vehicles, i.e. $(s_n^{t,d}, c_n^{t,d}, \hat{a}_1^{t,d}, \dots, \hat{a}_{n-1}^{t,d})$, which leads to substantially larger input dimension.
\end{remark}
\color{black}

\section{Deep Reinforcement Learning With Aggregated States} \label{sec:deep-rl}
The adoption of atomic actions has significantly reduced the {policy output} dimension. However, computing the optimal atomic action policy is still challenging due to the large state space, which scales significantly with the fleet size, number of regions, and battery discretization. In this section, we provide an efficient algorithm to train the fleet dispatching policy by incorporating our atomic action decomposition into the Proximal Policy Optimization (PPO) framework \citep{schulman2017proximal}. To tackle with the large state size, we use neural networks to approximate both the value function and the policy function, to be specified later. We also further reduce the state size using the following state reduction scheme:

\paragraph{Battery Level Clustering.} We map the state representation of all vehicle statuses into vehicle statuses with aggregated battery levels. We cluster the battery levels into 3 intervals, each of which denotes low battery level $\battery_L$, medium battery level $\battery_M$, and high battery level $\battery_H$, respectively. It is also possible to cluster the battery levels differently. If computing resources allow, we can cluster the battery levels with finer granularity, e.g. into 10 levels instead of 3. {We note that the cutoff points need not be chosen as uniform partitions of the $B$ battery levels. Instead, they can be selected according to thresholds in the nonlinear charging curve, corresponding to regions where the charging speed changes significantly.}

\paragraph{Trip Order Status Clustering.} In the state reduction scheme, trip orders are aggregated by recording only the number of requests originating from or arriving at each region, instead of tracking the number of trip requests for each origin-destination pair.
The clustering of trip order statuses reduces the state dimension from $O(\vert \Region \vert^2)$ to $O(\vert \Region \vert)$. 

{We emphasize that this state aggregation is {used for the neural network inputs}, while the full state vector is retained for simulation and policy evaluation. Although the aggregation inevitably removes some information about the exact battery status of vehicles and the detailed trip distribution, our numerical experiments show that the fleet dispatching policy learned under this reduced representation still achieves strong performance (see Section~\ref{sec:numerical}).}


We denote the state space after the reduction on the original state as $\bar{\mathcal{S}}$, which we refer to as the ``reduced state space". 
We use neural network $\hat{\pi}_{\theta}: \bar{\mathcal{S}} \times \Cartype \rightarrow \Delta(\hat{\mathcal{A}})$ on the reduced state space to approximate the atomic action policy function, where $\theta$ is the parameter vector for the atomic policy network.
\begin{algorithm}[h]
    \SetAlgoLined
    \caption{The Atomic-PPO Algorithm} \label{algo:ppo}
    \KwInputs{Number of policy iterations $M$, number of trajectories per policy iteration $K$, number of days per trajectory $D$, initial policy network $\hat{\pi}_{\theta_0}$.}
    \For{policy iteration $m = 1, \dots, M$}{
        Run policy $\hat{\pi}_{\theta_{m - 1}}$ over $K$ Monte Carlo trajectories, each spanning $\totaldays$ days and $\Horizon$ time steps per day. Collect the dataset as in \eqref{eq:ppo-data}.\\
        Compute empirical estimate of the long-run average daily reward \eqref{eq:ppo-g}.\\
        Compute empirical estimates of the relative value function \eqref{eq:ppo-v-mc}.\\
        Update the relative value network by minimizing the mean squared error between the network output and the empirical estimates in \eqref{eq:ppo-v-norm}.\\
        Estimate advantage functions as in \eqref{eq:advantage}.\\
        Obtain the updated policy network $\hat{\pi}_{\theta_m}$ by maximizing the surrogate objective function \eqref{eq:ppo-obj}.
    }
    \Return{policy $\hat{\pi}_{\theta_M}$}
\end{algorithm}
Our Atomic-PPO algorithm is formally presented in Algorithm \ref{algo:ppo}. For each policy iteration $m = 1, \dots, M$, we maintain a copy of the policy neural network parameters $\theta_{m-1}$ from the previous iteration and keep it fixed throughout the iteration. We then generate a dataset $\mathrm{Data}^{(K)}_{\theta_{m-1}}$ by simulating $K$ trajectories given the atomic action policy $\hat{\pi}_{\theta_{m-1}}$.
For each decision epoch $(t, d)$ and each trajectory $k$, this dataset includes the reduced state $\bar{s}^{t,d, (k)}_{n}$, the status $\cartype{n}^{t,d,(k)}$ of vehicle to be assigned, atomic action $\hat{a}^{t,d, (k)}_{n}$, and atomic reward $r^t(\cartype{n}^{t,d,(k)}, \hat{a}^{t,d, (k)}_{n})$ for each atomic step with $n=1, \dots, N$. That is, 
\begin{align} \label{eq:ppo-data}
    &\mathrm{Data}^{(K)}_{\theta_{m-1}} := \nonumber\\ 
    &\left\{\left[\left[\left(\bar{s}^{t, d, (k)}_{n}, \cartype{n}^{t,d,(k)}, \hat{a}^{t, d, (k)}_{n}, \right.\right.\right.\right. \notag\\
    &\quad\left.\left.\left.\left. r^t(\cartype{n}^{t,d,(k)}, \hat{a}^{t, d, (k)}_{n}) \right)_{n = 1}^{N} \right]_{t = 1}^{\Horizon} \right]_{d=1}^{\totaldays} \right\}_{k = 1}^K.
\end{align}
In each trajectory, we truncate the roll-out to $D$ days, with $T$ time steps in each day. Here, we set $D$ to be number that is large enough so that the distribution of the system state on day $D$ is close to the stationary state distribution regardless of the initial state. The procedure for the sequential assignment of atomic actions to individual vehicles follows from Section \ref{sec:atomic-action}.
Using the collected data, we estimate:\\
\underline{(i) The long-run average daily reward $g_{\theta_{m-1}} := R(\hat{\pi}_{\theta_{m-1}})$ for policy $\hat{\pi}_{\theta_{m-1}}$} \footnote{For simplicity, we present the algorithm under the assumption that the system has a single recurrent class, so the long-run average reward does not depend on the initial system state. Our approach can generalize to multi-chain systems.}:
\begin{equation} \label{eq:ppo-g}
    \hat{g}_{\theta_{m-1}} = \frac{1}{KD} \sum_{k = 1}^K \sum_{d = 1}^{\totaldays} \sum_{t = 1}^{\Horizon} \sum_{n = 1}^{\Size} r^t(\cartype{n}^{t,d,(k)}, \hat{a}^{t,d,(k)}_n),
\end{equation}
where we use $\hat{g}_{\theta_{m-1}}$ to denote the empirical estimate of long-run average daily reward of $\hat{\pi}_{\theta_{m-1}}$.\\
\underline{(ii) The relative value function $h_{n,\theta_{m-1}}: \mathbb{R}^{\vert \mathcal{S} \vert} \rightarrow \mathbb{R}$ given policy $\hat{\pi}_{\theta_{m-1}}$ for each atomic step $n \in [N]$}:\\
The relative value function quantifies how much better (or worse) each state is relative to the average. 
In particular, for any state $s \in \mathcal{S}$, where we recall that $s$ contains the time of day $t$ information, we define its relative value $h_{n, \theta_{m-1}}(s)$ under policy $\hat{\pi}_{\theta_{m-1}}$ at atomic step $n \in [N]$ as:
\begin{align} \label{eq:ppo-v-def}
    &h_{n, \theta_{m-1}}(s) \notag\\ 
    &= \mathbb{E}_{\hat{\pi}_{\theta_{m-1}}}\left[\sum_{i = n}^{N} \left(r^t(\cartype{i}^{t,1}, \hat{a}^{t,1}_{i}) - \frac{1}{TN}g_{\theta_{m-1}}\right) \right. \notag \\
    &+ \sum_{\ell = t + 1}^{\Horizon} \sum_{i = 1}^{N} \left( r^{\ell}(\cartype{i}^{\ell,1}, \hat{a}^{\ell,1}_{i}) - \frac{1}{TN}g_{\theta_{m-1}} \right) \notag\\
    &\qquad \left. \Bigg\lvert s^{t,1}_n = s \right] \notag \\
    &+ \sum_{d = 2}^{\infty} \sum_{\ell = 1}^{\Horizon} \sum_{i = 1}^{N} \notag\\
    & \mathbb{E}_{\hat{\pi}_{\theta_{m-1}}}\left[ \left( r^{\ell}(\cartype{i}^{\ell,d}, \hat{a}^{\ell,d}_{i}) - \frac{1}{TN}g_{\theta_{m-1}} \right) \right.\notag\\ 
    &\quad \left. \Bigg\lvert s^{t,1}_n = s \right], \quad \forall s \in \mathcal{S},\  \forall t \in [\Horizon].
\end{align}

\begin{remark}By Proposition 2 in \cite{dai2025optimal}, the infinite series in \eqref{eq:ppo-v-def} is well defined. Additionally, under the atomic action decomposition, the Markov chain has a period of $TN$, which is the total number of atomic steps in each day. The definition of the relative value function is equivalent to the one defined using the Cesaro limit, as given by Puterman (see page 338 of \cite{PutermanMDP}) for periodic chains, up to an additive constant (see Proposition 3 in \cite{dai2025optimal}).
\end{remark}

For any state $s^{t,d,(k)}_n$ in the atomic step $n$ of trajectory $k$ of decision epoch $(t, d)$, we construct an empirical estimate $\hat{h}^{t,d,(k)}_{n, \theta_{m-1}}$ of its relative value function $h_{n, \theta_{m-1}}(s^{t,d,(k)}_n)$ as:
\begin{align} \label{eq:ppo-v-mc}
    &\hat{h}^{t,d, (k)}_{n,\theta_{m-1}} \notag\\
    &:= \sum_{i = n}^{N} \left(r^t(\cartype{i}^{t,d,(k)}, \hat{a}^{t,d, (k)}_{i}) - \frac{1}{TN}\hat{g}_{\theta_{m-1}}\right) \notag \\
    &+ \sum_{\ell = t + 1}^{\Horizon} \sum_{i = 1}^{N} \left( r^{\ell}(\cartype{i}^{\ell,d,(k)}, \hat{a}^{\ell,d, (k)}_{i}) - \frac{1}{TN}\hat{g}_{\theta_{m-1}} \right) \notag \\
    &+ \sum_{d' = d + 1}^{\totaldays} \sum_{\ell = 1}^{\Horizon} \sum_{i = 1}^{N} \left( r^{\ell}(\cartype{i}^{\ell,d',(k)}, \hat{a}^{\ell,d', (k)}_{i}) - \frac{1}{TN}\hat{g}_{\theta_{m-1}} \right).
\end{align} 
Due to the large state space, we cluster the state into reduced states as described in earlier parts of Section \ref{sec:deep-rl}. Then, we use neural networks $h_{\psi_{m-1}}: \bar{\mathcal{S}} \rightarrow \mathbb{R}$ on the reduced state space to approximate the relative value function for all atomic steps, where $\psi_{m-1}$ is the network parameters. 
We learn $h_{\psi_{m-1}}$ by minimizing the mean-square loss given the empirical estimates:
\begin{equation} \label{eq:ppo-v-norm}
    \sum_{k = 1}^K \sum_{d = 1}^{\totaldays} \sum_{t = 1}^{\Horizon} \sum_{n = 1}^{\Size} \left( h_{\psi_{m-1}}(\bar{s}^{t,d, (k)}_n) - \hat{h}^{t,d, (k)}_{n,\theta_{m-1}}\right)^2.
\end{equation}
\underline{(iii) The advantage function of $\hat{\pi}_{\theta_{m-1}}$}:\\
The advantage function quantifies the change in the long-run average reward resulting from deviating from the previous stage policy $\hat{\pi}_{\theta_{m-1}}$ by selecting the atomic action $\hat{a}^{t,d, (k)}_{n}$ at a given state $s^{t,d, (k)}_{n}$.
The empirical estimates of the advantage function is given by:
\begin{align} \label{eq:advantage}
    &\hat{A}_{\theta_{m-1}}(\bar{s}^{t,d, (k)}_{n}, \cartype{n}^{t,d,(k)}, \hat{a}^{t,d, (k)}_{n}) := \nonumber\\ 
    &\quad \begin{cases}
        r^t(\cartype{n}^{t,d,(k)}, \hat{a}^{t,d, (k)}_{n}) - \frac{1}{TN}\hat{g}_{\theta_{m-1}} + \\
        h_{\psi_{m-1}}(\bar{s}^{t,d, (k)}_{n + 1}) - h_{\psi_{m-1}}(\bar{s}^{t,d, (k)}_{n}), \\
        \quad \text{for all } n < N,\\
        r^t(\cartype{n}^{t,d,(k)}, \hat{a}^{t,d, (k)}_{n}) - \frac{1}{TN}\hat{g}_{\theta_{m-1}} + \\
        h_{\psi_{m-1}}(\bar{s}^{t + 1,d, (k)}_{1}) - h_{\psi_{m-1}}(\bar{s}^{t,d, (k)}_{n}), \\
        \quad \text{if } n = N, t < \Horizon,\\
        r^t(\cartype{n}^{t,d,(k)}, \hat{a}^{t,d, (k)}_{n}) - \frac{1}{TN}\hat{g}_{\theta_{m-1}} + \\
        h_{\psi_{m-1}}(\bar{s}^{1,d+1, (k)}_{1}) - h_{\psi_{m-1}}(\bar{s}^{t,d, (k)}_{n}), \\
        \quad \text{if } n = N, t = \Horizon.\\
    \end{cases}
\end{align}


Using the estimated advantage function $\hat{A}_{\theta_{m-1}}$, the Atomic-PPO algorithm computes the atomic action policy function of the next iteration $\hat{\pi}_{\theta_m}$ by choosing parameter $\theta_m$ that maximizes the clipped surrogate function defined as follows:
\begin{align}
    &\hat{L}(\theta_m, \theta_{m-1}) \nonumber\\
    &:= \frac{1}{K}\sum_{k = 1}^K \sum_{d = 1}^{\totaldays} \sum_{t = 1}^{\Horizon} \sum_{n = 1}^{\Size} \nonumber \\
    &\ \min\left(\frac{\hat{\pi}_{\theta_m}(\hat{a}^{t,d, (k)}_{n} \vert \bar{s}^{t,d, (k)}_{n}, \cartype{n}^{t,d,(k)})}{\hat{\pi}_{\theta_{m-1}}(\hat{a}^{t,d, (k)}_{n} \vert \bar{s}^{t,d, (k)}_{n}, \cartype{n}^{t,d,(k)})} \cdot \right. \nonumber\\
    &\ \left. \hat{A}_{\theta_{m-1}}(\bar{s}^{t,d, (k)}_{n}, \cartype{n}^{t,d,(k)}, \hat{a}^{t,d, (k)}_{n}), \right. \nonumber \\
    &\ \left. \text{clip}\left(\frac{\hat{\pi}_{\theta_m}(\hat{a}^{t,d, (k)}_{n} \vert \bar{s}^{t,d, (k)}_{n},\cartype{n}^{t,d,(k)})}{\hat{\pi}_{\theta_{m-1}}(\hat{a}^{t,d, (k)}_{n} \vert \bar{s}^{t,d, (k)}_{n},\cartype{n}^{t,d,(k)})}, 1 - \epsilon, 1 + \epsilon \right) \cdot \right. \nonumber\\
    &\ \left. \hat{A}_{\theta_{m-1}}(\bar{s}^{t,d, (k)}_{n},\cartype{n}^{t,d,(k)}, \hat{a}^{t,d, (k)}_{n}) \right), \label{eq:ppo-obj}
\end{align}
where $\epsilon>0$ is a hyperparameter referred as the clip size of the training. The policy update builds on the original PPO that is known for its training stability by performing conservative policy updates. Specifically, the update uses the clipped surrogate function \eqref{eq:ppo-obj} to limit the change in the policy by bounding the ratio between the new and old policy probabilities within a clip range defined by $\epsilon$. 

Under Atomic-PPO, {the number of policy evaluations and stored observations per simulated day is $O(NT)$ for a fixed network and discretization. For fixed discretization parameters,} the dimension of the reduced state vector itself grows linearly with $\lvert V \rvert$. In practice, state transitions and neural network training can be efficiently implemented using vectorized operations (e.g., in PyTorch). As a result, for moderate values of $\lvert V \rvert$, the {observed total training time} remains largely insensitive to the number of regions (see our experiments on larger number of regions in Section \ref{sec:appendix-robustness}).
\color{black}

\section{Reward Upper Bound Provided By Fluid Approximation Model} \label{sec:fluid}
Before presenting the performance of our atomic PPO algorithm, we construct an upper bound on the optimal long-run average reward using fluid limit. This upper bound will be used to assess the performance of our atomic PPO algorithm in numerical experiments as shown in the next section. We formulate a fluid-based linear optimization program, where the fluid limit is attained as the fleet size approaches infinity, with both trip arriving rates and charger numbers scaling up proportionally to the fleet size. Under the fluid limit, the system becomes deterministic, and the fleet dispatching policy, which is a probability distribution of vehicle flows across all actions, reduces to a deterministic vector that represents the fraction of fleet assigned to each action at each time of the day.
We define the decision variables of the fluid-based optimization problem as follows:
\begin{enumerate}
    \item[-] \emph{Fraction of fleet for trip fulfilling $\lptripfulfill{}{} := \left(\lptripfulfill{\cartype{}, \triptype{}}{\time}\right)_{\cartype{} \in \Cartype, \triptype{} \in \Triptype, \time \in [T]}$}, where $\lptripfulfill{\cartype{}, \triptype{}}{\time}$ is the fraction of vehicles with status $\cartype{}$ fulfilling trip requests of status $\triptype{}$ at time $\time$.
    \item[-] \emph{Fraction of fleet for repositioning $\lpreroute{}{} := \left(\lpreroute{\cartype{}, \destination}{\time}\right)_{\cartype{} \in \Cartype, \destination \in \Region, \time \in [T]}$}, where $\lpreroute{\cartype{}, \destination}{\time}$ is the fraction of vehicles with status $\cartype{}$ repositioning to $\destination$ at time $\time$. 
    \item[-] \emph{Fraction of fleet for charging $\lpcharge{}{} := \left(\lpcharge{\cartype{}, \type}{\time}\right)_{\cartype{} \in \Cartype, \type \in \Type, \time \in [T]}$}, where $\lpcharge{\cartype{}, \type}{\time}$ is the fraction of vehicles with status $\cartype{}$ charging with rate $\type$ at time $\time$.
    \item[-] \emph{Fraction of fleet for continuing the current action $\lppass{}{} := \left(\lppass{\cartype{}}{\time}\right)_{\cartype{} \in \Cartype, \time \in [\Horizon]}$}, where $\lppass{\cartype{}}{\time}$ is the fraction of fleet with status $\cartype{}$ taking the passing action at time $\time$.
\end{enumerate}

The fluid-based linear program aims at maximizing the total reward achieved by the fluid policy:
{
\begin{align}
&\max_{\lptripfulfill{}{}, \lpreroute{}{}, \lpcharge{}{}, \lpslack{}{}, \lppass{}{}}  \Size \sum_{\time \in [\Horizon]}^{} \sum_{\cartype{} \in \Cartype} \left[\sum_{\origin \in \Region} \sum_{\destination \in \Region} \left[ \tripfulfillreward{,\origin\destination}{\time} \left(\sum_{\tripactivetime \in [\Lc]}^{} \lptripfulfill{\cartype{}, (\origin, \destination, \tripactivetime)}{\time} \right)  \right.\right. \notag\\
    &\qquad+ \left.\left. \reroutingreward{,\origin\destination}{\time} \lpreroute{\cartype{}, \destination}{\time} \right] + \sum_{\type \in \Type} \chargingreward{,\type}{\time} \lpcharge{\cartype{}, \type}{\time} \right], \label{eq:fluid_LP}\tag{Fluid-LP} \\
    &\text{s.t.} \quad \text{Constraints in Appendix }\ref{apx:fluid-original}.\notag
\end{align}
}

The constraints enforce flow conservation for each vehicle status, limit trip fulfillment by mean arrivals while respecting the connection patience time, and ensure that charger occupancy does not exceed capacity. They also require sufficient battery for trips and repositioning, fleet fractions that sum to one, and non-negative variables. Vehicle flows are periodic across days, linking the last time step of one day to the first time step of the next.

Let $\bar{R}$ be the optimal objective value of \eqref{eq:fluid_LP}. 

\begin{theorem} \label{thm:fluid-obj-val}
    $R^*(s) \leq \bar{R}, \quad \forall s \in \mathcal{S}$.
\end{theorem}

The proof of Theorem \ref{thm:fluid-obj-val} is provided in Appendix \ref{apx:fluid-proof}. Theorem \ref{thm:fluid-obj-val} shows that $\bar{R}$ is an upper bound on the long-run average daily reward that can be achieved by any feasible policy. In the numerical section, we assess the ratio between the average daily reward achieved by our Atomic-PPO and the fluid upper bound $\bar{R}$. This ratio provides a lower bound on the fraction of the optimal average daily reward attained by Atomic-PPO. 

\section{Numerical Experiments} \label{sec:numerical}
We conduct numerical experiments using the for-hire vehicle trip record data from July 2022, obtained from the New York City Taxi and Limousine Commission (TLC) \cite{nyctlc}. Each trip record includes the origin and destination taxi zones, base passenger fares, trip duration, and distance, all of which are used for model calibration.

\begin{figure}[tbp]
    \centering
    \begin{minipage}[t]{0.74\linewidth}
        \centering
        \begin{subfigure}[t]{0.48\linewidth}
            \centering
            \includegraphics[height=3in,width=\linewidth,keepaspectratio]{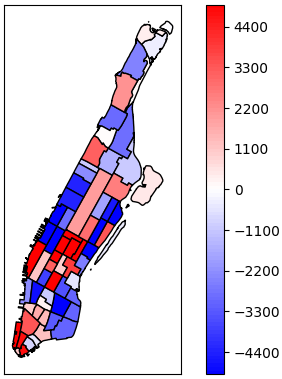}
            \caption{Weekday Morning Rush Hours\\\centering 7-10 am Mondays - Thursdays}
            \label{fig:manhattan-demand-morning-rush}
        \end{subfigure}\hfill
        \begin{subfigure}[t]{0.48\linewidth}
            \centering
            \includegraphics[height=3in,width=\linewidth,keepaspectratio]{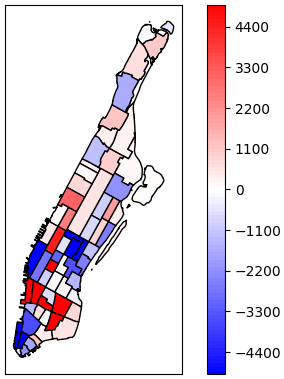}
            \caption{Weekday Evening Rush Hours\\\centering 5-8 pm Mondays - Thursdays}
            \label{fig:manhattan-demand-evening-rush}
        \end{subfigure}
        \caption{Manhattan Trip Demand Imbalance.}
        \label{fig:manhattan-demand}
    \end{minipage}\hfill
    \begin{minipage}[t]{0.23\linewidth}
        \centering
        \includegraphics[height=3in,width=\linewidth,keepaspectratio]{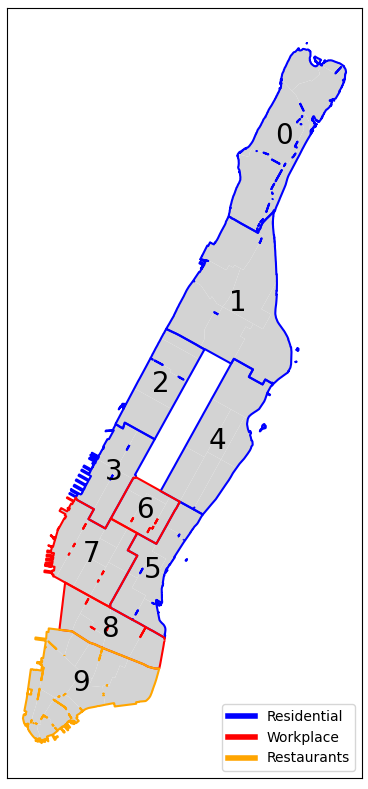}
        \caption[Service Regions in Manhattan.]{\raggedright Service Regions in Manhattan.}
        \label{fig:map}
    \end{minipage}
\end{figure}

We consider trip requests from 0:00 to 24:00, Mondays to Thursdays, when trip demand follows a similar pattern that differs from Fridays through Sundays. Each workday is partitioned into time intervals of 5 minutes. Ride-hailing trips in NYC exhibit significant demand imbalances during the morning (7--10 am) and evening (5--8 pm) rush hours. As shown in Figure \ref{fig:manhattan-demand}, certain zones experience more inbound trips (highlighted in red), while others see more outbound trips (in blue).
We partition Manhattan into 10 regions (Figure \ref{fig:map}) by aggregating adjacent taxi zones with similar demand patterns in both morning and evening rush hours. Based on trip request patterns, we identify 3 categories of regions: workplaces (red), restaurant areas (orange), and residential areas (blue). 

Using the NYC TLC dataset, we build a simulator of trip order arrival with the environment parameters in Table \ref{tab:env-params}. We calibrate trip arrival rates, durations, and rewards by averaging the corresponding records for each origin-destination pair and hour of the day. We scale down mean trip arrivals by the ratio of our chosen fleet size (300) to the estimated actual fleet size (12.8k).

We calibrate battery consumption using the Chevrolet Bolt EUV model. Because computation and sensing for autonomous driving are estimated to consume half of the battery, we reduce its nominal range of 260 miles to an effective range of 130 miles. We use an outlet rate of 75 kW for DC fast chargers as the default charging rate unless otherwise specified \cite{afsl, evinfra}. We adopt the non-linear charging curve for charging (Table \ref{tab:nonlinear-charging-curve}). 

\begin{table}[htbp]
    \centering
    \begin{minipage}[t]{0.56\linewidth}
        \vspace{0pt}
        \centering
\begin{tabular}{|c|c|}
        \hline
        Parameter & Value \\
        \hline
        Decision Epoch Length & 5 mins\\
        Number of Time Steps Per Day & 288\\
        Number of Days Per Episode & 8\\
        Fleet Size & 300\\
        Battery Pack Size & 65 kWh\\
        Vehicle Range & 130 miles\\
        Initial Vehicle Battery & 50\% Charged\\
        Charger Outlet Rate & 75 kW\\
        \hline
    \end{tabular}
        \caption{Simulation environment parameters.}
        \label{tab:env-params}
    \end{minipage}\hfill
    \begin{minipage}[t]{0.40\linewidth}
        \vspace{0pt}
        \centering
        \begin{tabular}{|c|c|}
            \hline
            \shortstack{Battery\\Percentage} & \shortstack{Charging\\Time (s)} \\
            \hline
            0\%-10\% & 47 \\
            10\%-40\% & 33 \\
            40\%-60\% & 40 \\
            60\%-80\% & 60 \\
            80\%-90\% & 107 \\
            90\%-95\% & 173 \\
            95\%-100\% & 533 \\
            \hline
        \end{tabular}
        \caption{Charging time (seconds) for each percentage of battery increase.}
        \label{tab:nonlinear-charging-curve}
    \end{minipage}
\end{table}

We present the details about the simulation environment and the Atomic-PPO training procedure (Algorithm \ref{algo:ppo}) in Appendix \ref{sec:appendix-training}. With 30 CPUs running in parallel, training a policy for 300 vehicles takes 4 hours, with approximately 10 minutes per policy iteration.


We use the power-of-k dispatch policy and the fluid-based policy as our benchmark algorithms. The power-of-$k$ dispatch policy considers trip requests in revenue-descending order. For each trip request, it selects the $k$ vehicles with the smallest remaining travel time to the trip origin region and assigns the vehicle with the highest battery level. The assigned vehicle must be no more than $\Lp$ time steps away from the trip origin and have sufficient battery to complete the trip and then drive to the nearest charger. Requests for which no vehicle satisfies these requirements are queued or abandoned. After trip assignments, idle vehicles below full charge are assigned to the nearest charger to charge for $\chargetime$ time steps. All other vehicles remain idle or continue their current tasks. We experimented with $k=1,\ldots,5$ and found that $k=2$ achieved the highest average reward, so we use $k=2$ throughout. The asymptotic service-level guarantee in \cite{varma2023electric}, which assumes spatially uniform demand and charger locations, does not imply reward optimality in our spatially heterogeneous setting.

The fluid policy is constructed from the optimal solution of \eqref{eq:fluid_LP}. For each pair $(v,\eta)$, we aggregate, separately for each action, the LP flows of vehicles with destination region $v$ and remaining time $\eta$ across all battery levels, and normalize them to obtain the corresponding proportions assigned to each action. At each decision epoch, these proportions are applied to the realized number of vehicles in each status $(v,\eta,b)$ to obtain the target number of vehicles of that status assigned to each action. We then convert the resulting fractional assignments into integer vehicle assignments using an adaptation of the bipartite dependent-rounding procedure in \citet{gandhi2006dependent}.


In Section \ref{subsec:superior}, we compare Atomic-PPO with the fluid based reward upper bound $\bar{R}$ and the benchmark policies; Section \ref{subsec:charger-deployment} examines charger deployment and fleet acquisition. Appendix \ref{sec:appendix-robustness} presents robustness checks under demand distribution shifts, stochastic trip durations, and a larger number of regions.


\subsection{Experiments with Abundant Chargers} \label{subsec:superior}
In this subsection, we evaluate the performance of Atomic-PPO against benchmark policies in a setting with abundant (3,000) chargers, which removes charging availability as a limiting factor and isolates the algorithm’s ability to make effective dispatching decisions.

Table \ref{tab:payoff} shows that Atomic-PPO outperforms both the fluid policy and the power-of-$k$ policy by a large margin. The long-run average reward attained by Atomic-PPO is 91\% of the fluid upper bound. In contrast, the power-of-$k$ policy achieves 81\%, while the fluid policy reaches 69\% of the fluid upper bound. Table \ref{tab:fleet-trip-periods} reports the fraction of the fleet serving trips, the fraction of requests fulfilled, and the average hourly reward during morning rush hours, evening rush hours, and non-peak hours. Atomic-PPO fulfills a higher fraction of requests and achieves a higher average hourly reward than the fluid policy and power-of-$k$ in all three periods, with the largest advantage during morning rush hours. It also maintains the highest fraction of the fleet assigned to trip fulfillment during morning rush and non-peak hours and achieves close to full fleet utilization for trip fulfillment during evening rush hours.

\begin{table}[h]
    \centering
    \begin{tabular}{c|c|c}
        Algorithm & Reward/$\bar{R}$ & Avg. Daily Reward \\
        \hline
        Atomic-PPO & 91\% & \$390K\\
        Atomic-PPO (without repositioning) & 89\% & \$385K\\
        Power of k & 81\% & \$345K\\
        Fluid policy & 69\% & \$295K\\
    \end{tabular}
    \caption{Reward achieved by Atomic-PPO with and without repositioning, power-of-k policy and fluid policy.}    \label{tab:payoff}
\end{table}

\begin{table}[htb]
    \centering
    \setlength{\tabcolsep}{4pt}
    \renewcommand{\arraystretch}{1.1}
    \begin{tabular}{@{}llrrr@{}}
        \hline
        Period & Policy
        & \shortstack{Fleet Serving Trips (\%)}
        & \shortstack{Trips Fulfilled (\%)}
        & \shortstack{Reward (\$/Hour)} \\
        \hline
        Morning rush & Atomic-PPO & 82.7\% & 58.1\% & \$20K \\
        Morning rush & \shortstack[l]{Atomic-PPO (without\\repositioning)} & 76.7\% & 55.1\% & \$18K \\
        Morning rush & Power-of-$k$ & 63.3\% & 37.6\% & \$15K \\
        Morning rush & Fluid policy & 47.8\% & 36.1\% & \$12K \\
        \hline
        Evening rush & Atomic-PPO & 92.2\% & 46.4\% & \$22K \\
        Evening rush & \shortstack[l]{Atomic-PPO (without\\repositioning)} & 96.3\% & 49.9\% & \$22K \\
        Evening rush & Power-of-$k$ & 99.8\% & 42.7\% & \$22K \\
        Evening rush & Fluid policy & 75.9\% & 44.7\% & \$19K \\
        \hline
        Non-peak & Atomic-PPO & 65.1\% & 69.1\% & \$15K \\
        Non-peak & \shortstack[l]{Atomic-PPO (without\\repositioning)} & 64.5\% & 69.0\% & \$15K \\
        Non-peak & Power-of-$k$ & 58.2\% & 53.8\% & \$13K \\
        Non-peak & Fluid policy & 46.9\% & 53.2\% & \$11K \\
        \hline
    \end{tabular}
    \caption{Fleet serving trips, trips fulfilled, and mean reward per hour by time of day.}
    \label{tab:fleet-trip-periods}
\end{table}

The fluid policy is state-independent and does not account for the stochasticity of trip demand. Consequently, its performance can be limited when the fleet size is moderate and stochastic fluctuations remain significant relative to the system scale. In fact, when we scale the fleet size up to 1,200 vehicles, the fluid policy achieves 81\% of the fluid upper bound, suggesting that its performance improves as the system approaches the fluid limit.

Power-of-$k$ assigns trip requests based on a pre-specified order and does not account for the future spatial distribution of the fleet and trip demand. In contrast, Atomic-PPO can shape the future fleet distribution both through vehicle repositioning and through its selection of trip requests to fulfill. To isolate the effect of vehicle repositioning, which is absent from power-of-$k$, we additionally conduct an experiment in which vehicle repositioning is disabled for Atomic-PPO throughout the day. Even under this restriction, Atomic-PPO achieves 89\% of the fluid upper bound, still exceeding power-of-$k$ by a large margin, while the percentage of the fleet serving trip requests, the percentage of requests fulfilled, and the average hourly reward remain comparable to those under Atomic-PPO with repositioning enabled across all periods. During evening rush hours, the no-repositioning variant has slightly higher percentages of the fleet serving trips and requests fulfilled, while achieving the same average hourly reward as Atomic-PPO with repositioning enabled. This comparison indicates that Atomic-PPO's primary advantage comes from selectively fulfilling trip requests rather than vehicle repositioning in this Manhattan experiment.

\begin{figure}[htb]
    \centering
    \includegraphics[width=\linewidth]{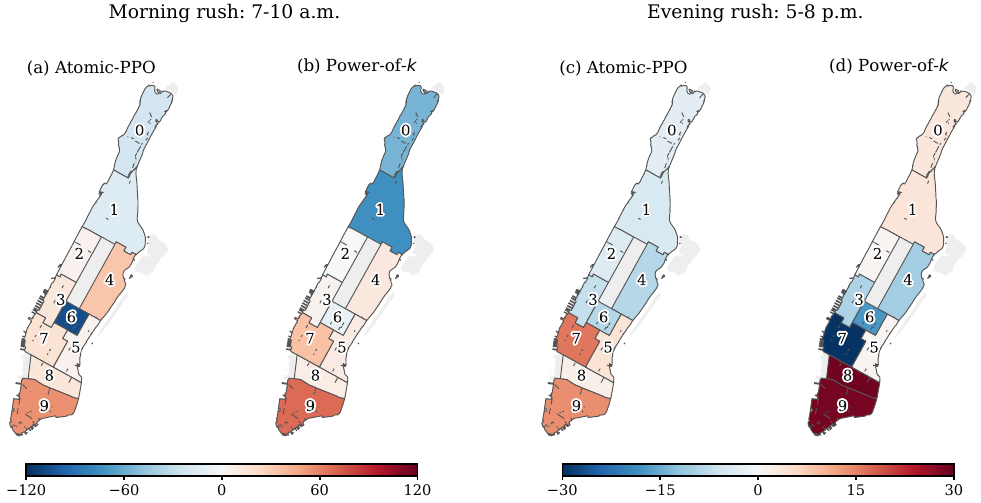}
    {
    \phantomsubcaption\label{fig:vehicle-flow-morning-rush-ppo}%
    \phantomsubcaption\label{fig:vehicle-flow-morning-rush-d-closest}%
    \phantomsubcaption\label{fig:vehicle-flow-evening-rush-ppo}%
    \phantomsubcaption\label{fig:vehicle-flow-evening-rush-d-closest}%
    }
    \caption{Net passenger-driven redistribution during weekday rush hours under Atomic-PPO with vehicle repositioning disabled and power-of-$k$ with $k=2$. Values are inbound minus outbound passenger-trip assignments per region. Red indicates net inflow and blue net outflow.}
    \label{fig:vehicle-flow-comparison}
\end{figure}

Figure \ref{fig:vehicle-flow-comparison} compares the resulting vehicle flows under power-of-$k$ and Atomic-PPO with vehicle repositioning disabled. Here, net passenger-driven redistribution is measured as inbound minus outbound vehicle assignments induced by passenger-trip fulfillment, summed over each rush-hour window. During the morning rush hours, Atomic-PPO generates a net outflow of approximately 106 vehicle assignments from Midtown region 6 (Figure \ref{fig:vehicle-flow-morning-rush-ppo}), compared with approximately 10 under power-of-$k$ (Figure \ref{fig:vehicle-flow-morning-rush-d-closest}). This outflow counteracts the tendency of morning passenger demand to concentrate vehicles in Midtown, as illustrated in Figure \ref{fig:manhattan-demand-morning-rush}. During the evening rush hours, Atomic-PPO produces a net inflow of approximately 2 vehicle assignments into region 8, compared with approximately 29 under power-of-$k$, thereby limiting the accumulation of vehicles in this area. Together with Figure \ref{fig:manhattan-demand}, these patterns suggest that Atomic-PPO uses trip fulfillment to redistribute vehicles and mitigate spatial supply-demand imbalances, even when vehicle repositioning is disabled.

\begin{remark}
    The modest additional benefit of repositioning in the Manhattan experiment should not be generalized to other settings. We conjecture that when demand across origin--destination pairs is sufficiently high, Atomic-PPO can rebalance the fleet largely through selective trip fulfillment, thereby reducing the need for empty-vehicle repositioning. 
\end{remark}

\subsection{Charger Deployment and Fleet Acquisition} \label{subsec:charger-deployment}
\subsubsection{Charger locations.}
We conduct experiments to study how charger deployments affect the average reward. Table \ref{tab:charging-facility-deployment} compares uniform deployments of 1--4 chargers per region with three non-uniform deployments of 15 chargers in Midtown, Upper, or Lower Manhattan and the abundant-charger setting in Section \ref{subsec:superior}.

\begin{table}[htb]
    \centering
    \setlength{\tabcolsep}{2pt}
    \begin{tabular}{|c|c|c|c|c|c|c|c|c|c|c|c|}
        \hline
         & \multicolumn{6}{c|}{Residential} & \multicolumn{3}{c|}{Workplace} & \multicolumn{1}{c|}{Restaurants} & \\
        \hline
        Region & 0 & 1 & 2 & 3 & 4 & 5 & 6 & 7 & 8 & 9 & Avg. Daily Reward\\
        \hline
        Unif. 10 chargers & 1 & 1 & 1 & 1 & 1 & 1 & 1 & 1 & 1 & 1 & \$265K\\
        Unif. 20 chargers & 2 & 2 & 2 & 2 & 2 & 2 & 2 & 2 & 2 & 2 & \$380K\\
        Unif. 30 chargers & 3 & 3 & 3 & 3 & 3 & 3 & 3 & 3 & 3 & 3 & \$385K\\
        Unif. 40 chargers & 4 & 4 & 4 & 4 & 4 & 4 & 4 & 4 & 4 & 4 & \$390K\\
        \hline
        Midtown Manhattan. & 0 & 0 & 0 & 0 & 0 & 0 & 0 & 15 & 0 & 0 & \$385K\\
        Upper Manhattan & 3 & 3 & 3 & 3 & 3 & 0 & 0 & 0 & 0 & 0 & \$280K\\
        Lower Manhattan & 0 & 0 & 0 & 0 & 0 & 3 & 3 & 3 & 3 & 3 & \$355K\\
        \hline
        Abundant Chargers & 300 & 300 & 300 & 300 & 300 & 300 & 300 & 300 & 300 & 300 & \$390K\\
        \hline
        Fluid Upper Bound & 300 & 300 & 300 & 300 & 300 & 300 & 300 & 300 & 300 & 300 & \$430K\\
        \hline
    \end{tabular}
    \caption{Average daily reward under different charging facility allocations.}
    \label{tab:charging-facility-deployment}
\end{table}

Table \ref{tab:charging-facility-deployment} shows that 40 uniformly deployed chargers achieve a reward comparable to the abundant-charger setting. With only 10 uniformly deployed chargers, the average reward drops to 68.3\% of the abundant-charger reward. In contrast, allocating all 15 chargers to Midtown Manhattan, in region 7, yields 98.6\% of the abundant-charger reward and a comparable average reward to 30 uniformly deployed chargers. As illustrated in Figure \ref{fig:manhattan-demand}, Midtown receives substantial passenger inflow during morning rush hours and generates substantial trip demand in the evening. Concentrating chargers in region 7 provides charging access near major passenger flows, which can limit the need for additional travel to reach chargers. This comparison highlights the benefit of locating chargers according to ridership patterns rather than distributing them uniformly.

\subsubsection{Charger quantity and outlet rates.}
For a fixed fleet size of 300 vehicles, we keep the chargers in Midtown Manhattan and consider deployments of 10, 15, and 20 units and charging outlet rates of 30 kW, 45 kW, 60 kW, and 75 kW. Figure \ref{fig:charger-deployment-vehnum=300} reports the average daily reward achieved by Atomic-PPO for each combination.

When the charging rate is relatively low (30 kW), increasing the charger number yields an approximately linear improvement in average daily reward. At 45 kW or higher, the gain from increasing the charger number from 10 to 15 exceeds that from 15 to 20. For any fixed charger number, increasing the charging rate exhibits diminishing returns, particularly at higher charger quantities.

These patterns suggest that the system transitions from being charging-capacity constrained to vehicle-availability constrained as charger quantity or outlet rate increases. Once charging constraints are largely mitigated, additional capacity has less value. Charger number and charger outlet rate act as partial substitutes over the configurations considered: higher outlet rates can compensate for fewer chargers, and vice versa, up to a moderate level of total charging capacity. Beyond this level, increasing either dimension in isolation yields limited benefit, suggesting that balanced investments across charger quantity and outlet rate are more effective than aggressively scaling one dimension alone.


\subsubsection{Impact of fleet size.}
\begin{figure}[htb]
    \centering
    \includegraphics[width=\linewidth]{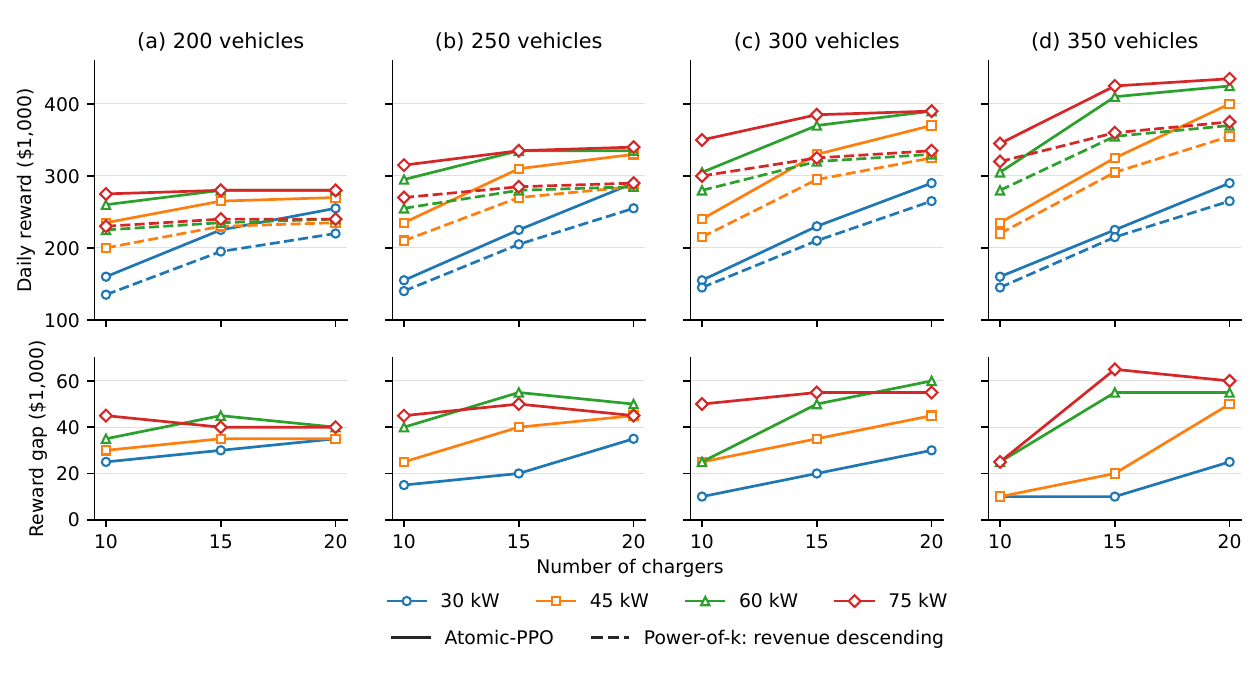}
    {
    \phantomsubcaption\label{fig:charger-deployment-vehnum=200}%
    \phantomsubcaption\label{fig:charger-deployment-vehnum=250}%
    \phantomsubcaption\label{fig:charger-deployment-vehnum=300}%
    \phantomsubcaption\label{fig:charger-deployment-vehnum=350}%
    }
    \caption{Average daily rewards (top) and reward gaps (Atomic-PPO minus power-of-$k$; bottom) across fleet sizes, charger quantities, and outlet rates. Solid lines show Atomic-PPO; dashed lines show power-of-$k$.}
    \label{fig:vehicle-charger-deployment}
\end{figure}

To study the joint impact of fleet size and charger deployment on average daily reward, we repeat the numerical experiments from the previous section under different fleet sizes: 200, 250, 300, and 350 vehicles. Trip arrival rates are held fixed at the baseline calibration while fleet size varies. Figure~\ref{fig:vehicle-charger-deployment} reports the average reward achieved by Atomic-PPO and power-of-$k$ for each combination of fleet size, charger quantity, and outlet rate. When charging capacity is limited, additional vehicles provide little improvement: with 10 chargers at 30 kW, rewards under Atomic-PPO remain between \$155K and \$160K across the four fleet sizes. With 20 chargers at 75 kW, in contrast, the reward increases from \$280K for 200 vehicles to \$435K for 350 vehicles. For the smaller fleet, rewards remain near \$280K once sufficient charging capacity is provided, whereas larger fleets benefit from additional charging capacity.
These findings suggest that the benefits of expanding either the fleet or charging capacity depend on the availability of the other, and hence that fleet expansion and charging deployment should be planned jointly, consistent with \cite{varma2023electric}. 

The gains from expanding charging capacity also depend on the dispatching policy. Across the same 48 configurations, Atomic-PPO achieves higher reported rewards than power-of-$k$, with daily reward gaps ranging from \$10K to \$65K. The gap generally increases with charging capacity: for 300 vehicles, it grows from \$10K with 10 chargers at 30 kW to \$55K with 20 chargers at 75 kW. Thus, Atomic-PPO extracts a larger reward gain from the same increase in charging capacity than power-of-$k$, indicating that the operational value of charging infrastructure depends in part on how the fleet is dispatched.

\color{black}

\subsubsection{Impact of vehicle range.} \label{subsec:range-outlet}
Finally, we return to the abundant-charger setting and double the original vehicle range (Table \ref{tab:charging-rate-range}). Both configurations achieve approximately \$390K per day, showing little change in the reported reward. The range experiment doubles the number of battery levels while holding per-mile battery consumption and charger outlet rate fixed; charging gains are recomputed using the same nonlinear charging curve. The limited gain is consistent with readily available chargers and the relatively short trips in Manhattan.

\begin{table}[htb]
    \centering
    \begin{tabular}{|c|c|c|c|}
        \hline
        Regime & Charger Power Rates & Vehicle Range & Avg. Daily Reward\\
        \hline
        Benchmark & 75 kW & 130 miles & \$390K\\
        Double Range & 75 kW & 260 miles & \$390K \\
        \hline
    \end{tabular}
    \caption{Impact of vehicle range on average reward under abundant chargers.}
    \label{tab:charging-rate-range}
\end{table}

\section{Concluding Remarks}
In this article, we develop a scalable Atomic-PPO algorithm for the efficient control of large-scale electric robo-taxi fleets, including trip matching, vehicle repositioning, and battery charging.
Our approach integrates atomic action decomposition into Proximal Policy Optimization (PPO), {using an atomic action space whose size is independent of fleet size} and significantly lowering the complexity of policy training.
Our state reduction technique substantially decreases the input dimension of the neural networks, improving runtime and memory usage. The Atomic-PPO algorithm makes it practical to compute near-optimal fleet dispatching policies under a wide range of charging infrastructure and fleet size configurations.

The computational efficiency of Atomic-PPO enables a large set of numerical experiments on how charger location, quantity, outlet rate, and fleet size jointly affect long-run system performance. Placing chargers according to observed ridership patterns yields higher system reward than distributing chargers uniformly across regions. For a fixed fleet size, increasing either charger quantity or outlet rate yields smaller reward gains when the other is already high. Larger fleets continue to benefit from additional charging capacity even when the same increase yields little improvement for smaller fleets, highlighting the need to plan fleet expansion and charging infrastructure together.

Future work includes incorporating heterogeneous rider patience and abandonment behavior, allowing vehicle repositioning to be interrupted by newly arriving trip requests, incorporating carpooling decisions into fleet control, and capturing more refined spatial and temporal granularity while preserving scalability.

\color{black}



\clearpage

\appendix
\section{Details of Simulation Environment and Atomic-PPO Training} 
\label{sec:appendix-training}
\subsection{Notations}\label{sec:appendix-notations}
\begin{table}[htb]
    \centering
    \begin{tabular}{|c|c|}
        \hline
        Symbol & Description \\
        \hline
        $\Region$ & The set of all service regions\\
        $\Size$ & Fleet Size \\
        $\range$ & Vehicle battery size\\
        $\Type$ & Set of charging rates\\
        $\Horizon$ & Number of time steps of each single day\\
        \hline
        $\timecost{\origin\destination}{\time}$ & Trip duration from $\origin$ to $\destination$ at time $\time$\\
        $\battery_{\origin\destination}$ & Battery consumption for traveling from region $\origin$ to $\destination$\\
        $\n{\region}{\type}$ & Number of chargers of rate $\type$ at region $\region$ \\
        \hline
        $\arrrate{\origin\destination}{\time}$ & Trip arrival rate from $\origin$ to $\destination$ at time $\time$\\
        $\arrnum{\origin\destination}{\time,\day}$ & Number of trip requests from $\origin$ to $\destination$ at time $\time$ of day $\day$\\
        $\Lp$ & Maximum pickup patience time\\
        $\Lc$ & Maximum connection patience time\\
        $\chargetime$ & Duration of a charging period\\
        \hline
        $\Cartype$ & The set of all vehicle statuses, with a generic vehicle status denoted as $\cartype{}$\\
        $\Triptype$ & The set of all trip statuses, with a generic trip status denoted as $\triptype{}$\\
        $\Plugtype$ & The set of all charger statuses, with a generic charger status denoted as $\plugtype{}$\\
        \hline
        $\state{}{\time, \day}$ & The state vector at time $\time$ of day $\day$\\
        $\jointaction{}{\time, \day}$ & The fleet action vector at time $\time$ of day $\day$\\
        $\jointtripfulfillaction{\cartype{}, \triptype{}}{\time, \day}$ & The number of vehicles of status $\cartype{}$ assigned to fulfill trips of status $\triptype{}$ at time $\time$ of day $\day$ \\
        $\jointreroutingaction{\cartype{}, \region}{\time, \day}$ & The number of vehicles of status $\cartype{}$ assigned to reposition to region $\region$ at time $\time$ of day $\day$ \\
        $\jointchargingaction{\cartype{}, \type}{\time, \day}$ & The number of vehicles of status $\cartype{}$ assigned to charge with rate $\type$ at time $\time$ of day $\day$ \\
        $\jointpassaction{\cartype{}}{\time, \day}$ & The number of vehicles of status $\cartype{}$ assigned to pass at time $\time$ of day $\day$ \\
        $\reward{}{\time}(a)$ & The reward given action $a$ at time $\time$ \\
        \hline
    \end{tabular}
    \caption{Notations for the electric robo-taxi system.}
    \label{tab:notations}
\end{table}

\subsection{System Parameters}
We consider a robo taxi ride hailing system operating continuously over a 24 hour horizon. The service area is partitioned into 10 regions (see Figure \ref{fig:map}), and the fleet consists of 300 vehicles. System level decisions are made every 5 minutes, resulting in 288 decision epochs per day. We assume a homogeneous fleet calibrated using the Chevrolet Bolt EUV model with a nominal driving range of 260 miles. Autonomous vehicles are estimated to consume roughly 50\% of their battery capacity on onboard computation and sensing, which effectively halves the usable driving range after a full charge. Accordingly, we discretize each vehicle's battery into 130 evenly spaced levels, where one battery level is depleted for every mile traveled.

We use 75 kW DC fast chargers as the default outlet rate in all experiments unless otherwise specified \citep{afsl, evinfra}. Each charging period lasts 15 minutes, and the number of battery levels gained is determined according to Table \ref{tab:nonlinear-charging-curve}. When chargers with different outlet rates are considered, the battery gain per charging period is adjusted by scaling the charging time in each battery percentage bucket of Table \ref{tab:nonlinear-charging-curve} according to the ratio between 75 kW and the corresponding charger outlet rate. The charging cost incurred is set to be \$0.39 per kWh consumed at all times of the day \citep{nycdot_ev_charging}.

The direct repositioning cost is set to zero. Repositioning consumes battery and vehicle time, with electricity costs incurred when the battery is recharged. The full-day comparison in Table \ref{tab:payoff} assesses performance without vehicle repositioning between regions, including overnight; travel to chargers within a vehicle's current region remains allowed.

We calibrate trip durations and trip rewards using the NYC TLC dataset \citep{nyctlc}. For each origin destination region pair and each hour of the day, we compute the average trip duration and average trip fare across all observed trips, and round the average duration to the nearest 5 minutes. At each 5 minute decision epoch, the simulator uses the trip duration and reward corresponding to the hour in which that epoch occurs. In Appendix \ref{sec:appendix-robustness}, we conduct additional experiments using stochastic trip durations sampled from the empirical distribution. For each origin destination region pair and each hour of the day, we calibrate this empirical distribution by first rounding all observed trip durations to the nearest 5 minutes and then computing the fraction of trips in each 5 minute bin. During simulation, the trip duration is sampled according to these fractions.

For the trip demand profile, we first compute the average number of trips for each origin destination pair at each hour of the day, averaged over all Mondays to Thursdays in July 2022. We then scale these average arrivals by the ratio between the chosen fleet size (300 vehicles) and the estimated actual fleet size (12.8k vehicles), where the latter is inferred from the maximum number of simultaneous trips observed throughout the day. At each decision epoch, the simulator generates trip requests for every origin destination pair by sampling from a Poisson distribution whose mean equals the scaled average arrival rate corresponding to that pair and the hour in which the decision epoch occurs. 

\subsection{Atomic-PPO Training Procedure} \label{sec:appendix-dnn}
We use deep neural networks to approximate the policy and value functions. Both networks are implemented as three-layer feed forward architectures. The numbers of neurons in the hidden layers are 500, 64, and 64 for both the policy and value networks. The activation functions for the policy network are tanh, tanh, and tanh, while the value network uses tanh, relu, and tanh. The network input consists of the reduced state vector (defined in Section \ref{sec:deep-rl}) together with an embedding of the time of day. The time embedding is obtained by passing the time index through an embedding layer with dimension 24.

The initial clipping rate $\epsilon_0$ of Atomic-PPO is set to $0.1$, with an exponential decay rate $\gamma = 0.97$ across policy iterations. Specifically, the clipping rate at policy iteration $m = 1,2,\dots$ is given by $\max\{\epsilon_0 \cdot \gamma^m, 0.01\}$. In each policy iteration, we roll out 30 independent trajectories, each spanning 8 operation days, with vehicles initialized at half charge and uniformly distributed across all regions. Using the collected trajectories, we update the value network and the policy network according to Algorithm \ref{algo:ppo}. Both networks are trained using mini-batch updates. At each gradient step, we compute gradients from a batch of 1024 data points randomly sampled from the collected dataset and update the network parameters using Adam. The learning rates are set to $3 \times 10^{-4}$ for the value network and $5 \times 10^{-4}$ for the policy network. The numbers of gradient updates are 100 for the value network and 20 for the policy network. Table \ref{tab:training-params} summarizes all hyperparameters used in Atomic-PPO training.
 
\begin{table}[h]
    \centering
    \begin{tabular}{|c|c|c|}
        \hline
        Component & Hyperparameter Description & Value \\
        \hline
        \multirow{4}{*}{Training setup} 
        & Number of rollout trajectories per policy iteration & 30 \\
        & Number of operation days per trajectory & 8 \\
        & Initial PPO clipping rate & 0.1 \\
        & Exponential decay factor for clipping rate & 0.97 \\
        \hline
        \multirow{4}{*}{Policy network}
        & Time embedding dimension & 24 \\
        & Number of hidden layers & 3 \\
        & Neurons in each layer & [500, 64, 64] \\
        & Activation functions & [Tanh, Tanh, Tanh] \\
        & Learning rate (Adam optimizer) & $5\times10^{-4}$ \\
        & Mini-batch size per gradient update & 1024 \\
        & Number of gradient updates per iteration & 20 \\
        \hline
        \multirow{3}{*}{Value network}
        & Time embedding dimension & 24 \\
        & Number of hidden layers & 3 \\
        & Neurons in each layer & [500, 64, 64] \\
        & Activation functions & [Tanh, ReLU, Tanh] \\
        & Learning rate (Adam optimizer) & $3\times10^{-4}$ \\
        & Mini-batch size per gradient update & 1024 \\
        & Number of gradient updates per iteration & 100 \\
        \hline
    \end{tabular}
    \caption{Hyperparameters used for training Atomic-PPO.}
    \label{tab:training-params}
\end{table}

\color{black}



\section{Robustness Check of Atomic-PPO} \label{sec:appendix-robustness}

We evaluate the robustness of Atomic-PPO along three complementary dimensions that reflect realistic sources of model mismatch in ride-hailing systems: temporal shifts in trip demand profiles, stochasticity in trip durations, and increased state–action complexity induced by finer spatial partitions. Together, these experiments assess whether Atomic-PPO generalizes beyond the specific modeling assumptions and training instances used in the main experiments.

\paragraph{Distribution shift of trip demand profile.}
To assess the generalizability of Atomic-PPO under demand distribution shifts, we compare the long-run average daily reward achieved by Atomic-PPO trained on July’s demand profile with that achieved by Atomic-PPO retrained on each corresponding month’s demand profile, for June, August, and September (see Table \ref{tab:robustness-dist-shift}). This experiment reflects realistic non-stationarity in ride-hailing demand arising from seasonal variation, changes in travel patterns, and external factors. 

Specifically, we first construct the demand profile for July 2022 following the procedure described in Section \ref{sec:numerical}. This involves computing the trip demand volume for each origin–destination pair at each time of day, averaged across all workdays in July 2022, and then uniformly scaling the demand volumes according to the ratio between the chosen fleet size (300 vehicles) and the actual fleet size in Manhattan, defined as the maximum number of vehicles simultaneously serving trips over a day. We then construct the demand profiles for June, August, and September by applying the same uniform scaling factor to the corresponding average demand volumes for each origin–destination pair at each time of day.

\enlargethispage{\baselineskip}
Table \ref{tab:robustness-dist-shift} shows that Atomic-PPO trained on July’s demand profile achieves nearly identical long-run average daily reward to Atomic-PPO trained on the corresponding month’s demand profile across all three months, attaining as high as 91\% of the fluid-based upper bound for June and September and 95\% for August. These results indicate that Atomic-PPO maintains strong performance under month-to-month demand shifts, even without retraining on the updated demand profile.
\begin{table}[htb]
    \centering
    \begin{tabular}{lcccc}
        \toprule
         & \multicolumn{2}{c}{Trained on July's Demand} 
         & \multicolumn{2}{c}{Trained on Corresponding Month's Demand} \\
        \cmidrule(lr){2-3} \cmidrule(lr){4-5}
         & Avg.\ Daily Revenue & \% Fluid UB
         & Avg.\ Daily Revenue & \% Fluid UB \\
        \midrule
        June       & \$425K & 91\% & \$425K & 91\% \\
        August     & \$380K & 95\% & \$385K & 95\% \\
        September  & \$455K & 91\% & \$455K & 91\% \\
        \bottomrule
    \end{tabular}
    \caption{Comparison of Atomic-PPO performance when trained on July’s demand profile versus retrained on the corresponding month’s demand profile, reported separately for each month.}
    \label{tab:robustness-dist-shift}
\end{table}

\paragraph{Stochastic trip durations.} While in Section \ref{sec:numerical} we model trip durations for each origin–destination pair and time of day deterministically using their average values across workdays, real-world travel times are affected by congestion, incidents, and weather, introducing unavoidable stochasticity. To assess robustness to this source of uncertainty, we sample trip durations from the actual empirical distributions in the data.

Table \ref{tab:robustness-trip-duration} shows that Atomic-PPO achieves nearly identical long-run average daily reward under deterministic and stochastic trip durations. This suggests that the training and execution of Atomic-PPO are not sensitive to additional stochasticity in trip durations.

\begin{table}[htb]
    \centering
    \begin{tabular}{lcc}
        \toprule
         & Avg. Daily Reward
         & \% Fluid UB \\
        \midrule
        Deterministic trip duration       & \$390K & 91\%  \\
        Stochastic trip duration    & \$390K & 91\% \\
        \bottomrule
    \end{tabular}
    \caption{Performance comparison of Atomic-PPO under deterministic v.s. stochastic trip durations}
    \label{tab:robustness-trip-duration}
    \par\smallskip\begin{minipage}{0.95\linewidth}\footnotesize
    The deterministic-duration baseline uses the abundant-charger result from the first 20 training iterations.
    \end{minipage}
\end{table}

\paragraph{Larger number of regions.} To assess how Atomic-PPO scales with the number of regions, we partition Manhattan into 10, 15, and 20 regions, retrain Atomic-PPO for each instance, and compute the corresponding fluid-based revenue upper bound when feasible. Increasing the number of regions substantially enlarges the state and action spaces, making this a stringent test of scalability.

Across all three instances, Atomic-PPO’s total training time remains similar (within 4 hours), while peak memory usage increases from approximately 200~GB to 300~GB and 400~GB as the number of regions grows from 10 to 15 and 20. In contrast, a naive implementation of the fluid-based LP solves the 10-region instance in under 2 hours, requires roughly 14 hours for 15 regions, and fails with an out-of-memory error for 20 regions (exceeding 1~TB).

Table \ref{tab:robustness-larger-regions} shows that for the 10- and 15-region settings, where the fluid LP remains tractable, Atomic-PPO attains 91\% and 90\% of the fluid upper bound, respectively. Although the fluid upper bound cannot be computed for the 20-region instance, Atomic-PPO continues to achieve stable performance, with only a moderate reduction in reward. These results highlight that Atomic-PPO can scale to finer spatial partitions while maintaining strong performance, even in regimes where fluid-based benchmarks become computationally infeasible.

\begin{table}[htb]
    \centering
    \begin{tabular}{lcc}
        \toprule
         & Avg. Daily Reward
         & \% Fluid UB \\
        \midrule
        10 Regions       & \$390K & 91\%  \\
        15 Regions    & \$380K & 90\% \\
        20 Regions    & \$355K & N/A \\
        \bottomrule
    \end{tabular}
    \caption{Performance of Atomic-PPO across different number of regions}
    \label{tab:robustness-larger-regions}
    \par\smallskip\begin{minipage}{0.95\linewidth}\footnotesize
    The 10-region baseline uses the abundant-charger result from the first 20 training iterations.
    \end{minipage}
\end{table}

Taken together, these robustness checks indicate that Atomic-PPO generalizes well across multiple sources of modeling mismatch, including demand non-stationarity, stochastic trip durations, and increased spatial granularity. 

\color{black}

\section{State Transitions}\label{apx:state-transitions}
This appendix gives the state transition equations for the MDP in Section \ref{subsec:MDP}, using the state and action definitions introduced there.

\paragraph{State Transitions.} At any time $\time$ of day $\day$, given any state $\state{}{}$ and action $\jointaction{}{} \in \mathcal{A}_{\state{}{}}$, we compute the vehicle state vector at time $\time+1$. 
For each vehicle status $\cartype{} := (\destination, \eta, \battery) \in \Cartype$, 
{\allowdisplaybreaks[0]
\begin{align}
    &\state{(\destination, \eta, \battery)}{\time + 1, \day} = \left(\sum_{\origin \in \Region} \sum_{\cartype{}'= (\origin, \timetoarrival', \battery') \in \Cartype} \sum_{ \triptype{}= (\origin, \destination, \xi') \in \Triptype} \right. \nonumber\\
    &\qquad \left. \jointtripfulfillaction{\cartype{}', \triptype{}}{\time,\day}\mathds{1}(\timetoarrival' + \timecost{\origin\destination}{\time} - 1 = \timetoarrival, \  \battery' - \batterycost{\origin \destination} = \battery) \right) \tag{i} \\
    &\qquad+ \left(\sum_{\origin \in \Region} \sum_{\cartype{}'= (\origin, 0, \battery') \in \Cartype} \jointreroutingaction{\cartype{}', \destination}{\time,\day} \cdot \right. \nonumber\\
    &\qquad \left. \mathds{1}(\timecost{\origin\destination}{\time} - 1 = \timetoarrival, \  \battery' - \batterycost{\origin\destination} = \battery) \right) \tag{ii} \\
    &\qquad+ \left[ \left(\sum_{\type \in \Type} \jointchargingaction{(\destination, 0, \battery + \batterycost{\destination \destination} - \type \chargetime), \type}{\time,\day} \cdot \right. \right. \nonumber \\
    &\qquad \left. \mathds{1}(\timetoarrival = \timecost{\destination \destination}{\time} + \chargetime - 1, \battery \geq \type \chargetime - \batterycost{\destination \destination}) \right) +  \nonumber\\
    &\qquad \left. \left(\sum_{\battery' > \battery - \type \chargetime + \batterycost{\destination \destination}} \sum_{\type \in \Type} \jointchargingaction{(\destination, 0, \battery'), \type}{\time,\day} \cdot \right. \right. \nonumber\\
    &\qquad \left. \left. \mathds{1}(\timetoarrival = \timecost{\destination \destination}{\time} + \chargetime - 1, \battery = \range) \right) \right] \tag{iii}\\
    &\qquad+ \underbrace{\jointpassaction{(\destination, \timetoarrival, \battery)}{\time,\day} \mathds{1}(\timetoarrival = 0)}_{(iv)} + \underbrace{\jointpassaction{(\destination, \timetoarrival+1, \battery)}{\time,\day} \mathds{1}(\timetoarrival > 0)}_{(v)}, \label{eq:setup-car-state-transition}
\end{align}
}
where (i) and (ii) correspond to vehicles assigned to new trip-fulfilling or repositioning actions with destination $v$, respectively. The destination, time to arrival, and battery level of these vehicles are updated based on the newly assigned trips. Term (iii) corresponds to the vehicles assigned to charge at time $\time$. {The vehicle will first drive to the charger, which requires $\timecost{\destination \destination}{\time}$ time steps and consumes $\batterycost{\destination \destination}$ units of battery.} The battery level of these vehicles will be increased by $\delta \chargetime$ at the end of the charging period. If the vehicle is charged to full, then it will remain idle for the rest of the charging period. Term (iv) corresponds to the idling vehicles that continue to remain idle, and (v) corresponds to the vehicles taking the passing action whose remaining time of completing the current action decreases by $1$ in the next time step. 

Moreover, the trip state at time $\time+1$ of day $\day$ is given by \eqref{eq:setup-trip-state-transition}. For each trip status, we subtract the number of trip orders that have been fulfilled at time $\time$, and we increment the active time by $1$ for trip orders that are still in the system. We abandon the trip orders that have been active for more than $\Lc$ time steps. Additionally, new trip orders arrive in the system and we set their active time in the system to be $0$. That is, for all $(\origin, \destination) \in \Region \times \Region$,
\begin{align}\label{eq:setup-trip-state-transition}
     \state{(\origin, \destination, \xi)}{\time + 1, \day} = \left\{
     \begin{array}{ll}
                  \min\left\{\arrnum{\origin\destination}{\time, d}, \  \Size (\Lc + 1) \right\},  \\
                  \quad \text{if $\tripactivetime=0$}, \\
                  \state{(\origin, \destination, \tripactivetime - 1)}{\time,\day} - \sum_{\cartype{}  \in \Cartype} \jointtripfulfillaction{\cartype{}, (\origin, \destination, \tripactivetime-1)}{\time,\day}, \\
                  \quad \text{if $1 \leq \tripactivetime \leq \Lc$}.  
     \end{array}
     \right.
\end{align}
As mentioned in Section \ref{subsec:MDP}, we cap the number of new trip arrivals by $\Size(\Lc+1)$ so that the state space is finite. Then, trips that are not fulfilled at time $\time$ are queued to $\time+1$. Thus, the number of trips that have been in the system for $\tripactivetime \geq 1$ at time $\time+1$ equals to that with $\tripactivetime-1$ from time $\time$ minus the ones that are assigned to a vehicle at time $\time$.  

Lastly, the charger state at time $\time + 1$ of day $\day$ is given by \eqref{eq:setup-plug-state-transition}. For each charger status $\plugtype{} := (\region, \type, \plugremaintime) \in \Plugtype$: For all region $\region \in \Region$ and for all charger outlet rates $\type \in \Type$,
\begin{align}\label{eq:setup-plug-state-transition}
     \state{(\region, \type, \plugremaintime)}{\time + 1, \day} = \left\{
     \begin{array}{ll}
                  \sum_{\cartype{} = (\region, 0, \battery) \in \Cartype} \jointchargingaction{\cartype{}, \type}{\time,\day},  & \text{if } \plugremaintime=\chargetime - 1, \\
                  \state{(\region, \type, \plugremaintime + 1)}{\time, \day}, & \text{if } 0 < \plugremaintime< \chargetime - 1,\\
                  \state{(\region, \type, 1)}{\time, \day} + \left(\state{(\region, \type, 0)}{\time, \day}\right. \\
                  \left.- \sum_{\cartype{} = (\region, 0, \battery) \in \Cartype} \jointchargingaction{\cartype{}, \type}{\time,\day}\right), & \text{if } \plugremaintime=0.
     \end{array}
     \right.
\end{align}
At time $t+1$, the number of chargers that are occupied with remaining time $\plugremaintime = \chargetime - 1$ equals the total number of vehicles just assigned to charge at time $\time$. Chargers already in use at time $\time$ will have their remaining charging time $j$ decrease by 1 when the system transitions to time $\time + 1$. The number of chargers available (i.e. $j=0$) at time $\time + 1$ consists of (i) the chargers that have just completed their charging periods (i.e. $\state{(\region, \type, 1)}{\time, \day}$), and (ii) the chargers available at time $\time$ minus the ones that are assigned to charge vehicles (i.e. $\state{(\region, \type, 0)}{\time, \day} - \sum_{\cartype{} = (\region, 0, \battery) \in \Cartype} \jointchargingaction{\cartype{}, \type}{\time, \day}$).

\section{Fluid Based Linear Program}\label{apx:fluid-formulations}
\subsection{Original Formulation}\label{apx:fluid-original}
The original fluid-based linear program \eqref{eq:fluid_LP} uses the fleet fractions defined in Section \ref{sec:fluid} and the state transitions in Appendix \ref{apx:state-transitions}.

The constraints \eqref{eq:fluid-ev-conservation-red}-\eqref{eq:fluid-nonneg-red} are given as follows:
\begin{enumerate}
    \item The flow conservation for each vehicle status $\cartype{}:= (\destination, \timetoarrival, \battery) \in \Cartype$ at each time $\time$ of a day. \\
    In particular, the left-hand side of constraint \eqref{eq:fluid-ev-conservation-red} represents the fraction of fleet that transitions to vehicle status $\cartype{}$ according to \eqref{eq:setup-car-state-transition} at time $t$. The right-hand side of constraint \eqref{eq:fluid-ev-conservation-red} represents the total fraction of fleet with status $\cartype{}$ that is assigned to trip-fulfilling, repositioning, charging, and passing actions. 
    \begin{align}
        &\left(\sum_{\origin \in \Region} \sum_{\cartype{}'= (\origin, \timetoarrival', \battery') \in \Cartype} \sum_{ \triptype{}= (\origin, \destination, \xi') \in \Triptype} \right. \notag\\
        &\qquad \left. \lptripfulfill{\cartype{}', \triptype{}}{\time-1}\mathds{1}(\timetoarrival' + \timecost{\origin\destination}{\time-1} - 1 = \timetoarrival, \ \battery' - \batterycost{\origin \destination} = \battery) \right) \notag \\
        &\qquad+ \left(\sum_{\origin \in \Region} \sum_{\cartype{}'= (\origin, 0, \battery') \in \Cartype} \lpreroute{\cartype{}', \destination}{\time-1} \cdot \right. \notag\\
        &\qquad \left. \mathds{1}(\timecost{\origin\destination}{\time-1} - 1 = \timetoarrival, \  \battery' - \batterycost{\origin\destination} = \battery) \right) \notag \\
        &\qquad+ \left[ \left(\sum_{\type \in \Type} \lpcharge{(\destination, 0, \battery - \type \chargetime + \batterycost{\destination \destination}), \type}{\time-1} \cdot \right. \right. \notag\\
        &\qquad \left. \left. \mathds{1}(\timetoarrival = \timecost{\destination \destination}{\time} + \chargetime - 1, \battery \geq \type \chargetime - \batterycost{\destination \destination}) \right) + \right. \notag\\
        &\qquad \left. \left(\sum_{\battery' > \battery - \type \chargetime + \batterycost{\destination \destination}} \sum_{\type \in \Type} \lpcharge{(\destination, 0, \battery'), \type}{\time-1} \cdot \right. \right. \notag\\ 
        &\qquad \left. \left. \mathds{1}(\timetoarrival = \timecost{\destination \destination}{\time} + \chargetime - 1, \battery = \range) \right) \right] \notag\\
        &\qquad+ \lppass{(\destination, \timetoarrival, \battery)}{\time-1} \mathds{1}(\timetoarrival = 0) + \lppass{(\destination, \timetoarrival+1, \battery)}{\time-1} \mathds{1}(\timetoarrival < \maxtimecost{}) \notag \\
        &= \sum_{\triptype{} \in \Triptype} \lptripfulfill{\cartype{}, \triptype{}}{\time} + \sum_{\destination \in \Region} \lpreroute{\cartype{}, \destination}{\time} + \sum_{\type \in \Type} \lpcharge{\cartype{}, \type}{\time} + \lppass{\cartype{}}{\time},\notag\\ 
        &\qquad \forall \cartype{}:= (\destination, \timetoarrival, \battery) \in \Cartype, \   \time \in [\Horizon], \label{eq:fluid-ev-conservation-red}
    \end{align}
    We note that the time steps are periodic across days, and thus for $t=1$ in \eqref{eq:fluid-ev-conservation-red}, $t-1$ is the last time step $T$ of the previous day. Similarly, in all of the subsequent constraints \eqref{eq:fluid-passenger-flow-cap-red} -- \eqref{eq:fluid-charging-cap-red}, the time step $t$ on the superscript of a variable being negative indicates time step $T-t$ of the previous day and $t>T$ indicates time step $t-T$ of the next day. 
    \item Trip order fulfillment does not exceed trip order arrivals:
    \begin{align}
        &\sum_{\cartype{}=(\origin, \timetoarrival, \battery) \in \Cartype}^{}\sum_{\triptype{}=(\origin, \destination, \tripactivetime) \in \Triptype}^{} \lptripfulfill{\cartype{}, \triptype{}}{\time + \tripactivetime} \leq \frac{1}{\Size} \arrrate{\origin \destination}{\time}, \notag\\ 
        &\qquad \forall \origin, \destination \in \Region,\ \time \in [\Horizon].\label{eq:fluid-passenger-flow-cap-red}
    \end{align}
    \item The flow of charging vehicles in any region does not exceed the charger capacity.
    \begin{align}
        \sum_{j \in [\chargetime]} \sum_{\cartype{}= (\region, 0, \battery) \in \Cartype} \lpcharge{\cartype{}, \type}{\time - j} \leq \frac{1}{\Size}\n{\region}{\type}, \quad \forall \type \in \Type, \  \time \in [\Horizon]. \label{eq:fluid-charging-cap-red}
    \end{align}
    \item Vehicles have sufficient battery to fulfill trips:
    \begin{align}
        &\ \lptripfulfill{\cartype{}, \triptype{}}{\time} \mathds{1}\{\battery < \batterycost{\origin \destination} + \batterycost{\destination \destination}\} = 0,\notag\\ 
        &\quad \forall \cartype{}= (\origin, \timetoarrival, \battery) \in \Cartype,\ \triptype{}= (\origin, \destination, \tripactivetime) \in \Triptype, \ \time \in [\Horizon]. \label{eq:fluid-passenger-flow-battery-sufficiency-red}
    \end{align}
    \item Vehicles have sufficient battery to reposition:
    \begin{align}
        &\ \lpreroute{\cartype{}, \destination}{\time} \mathds{1}\{\battery < \batterycost{\origin \destination} + \batterycost{\destination \destination}\} = 0, \notag\\
        &\quad \forall \cartype{}= (\origin, \timetoarrival, \battery) \in \Cartype,\ \destination \in \Region, \ \time \in [\Horizon]. \label{eq:fluid-rerouting-flow-battery-sufficiency-red}
    \end{align}
    \item The fractions of fleet of all statuses sum up to 1 at all time steps:
    \begin{align}
        &\ \sum_{\cartype{} \in \Cartype} \left[ \sum_{\triptype{} \in \Triptype} \lptripfulfill{\cartype{}, \triptype{}}{\time} + \sum_{\destination \in \Region} \lpreroute{\cartype{}, \destination}{\time} + \sum_{\type \in \Type} \lpcharge{\cartype{}, \type}{\time} + \lppass{\cartype{}}{\time} \right] = 1,\notag \\
        & \qquad\qquad \qquad\qquad \qquad\qquad  \forall \time \in [\Horizon]. \label{eq:fluid-total-flow-red}
    \end{align}
    \item All variables are non-negative.
    \begin{align}
        \lptripfulfill{}{}, \lpreroute{}{}, \lpcharge{}{}, \lppass{}{} \geq 0. \label{eq:fluid-nonneg-red}
    \end{align}
\end{enumerate}

\subsection{Proof of Theorem \ref{thm:fluid-obj-val}}\label{apx:fluid-proof}
    By Theorem 9.1.8 on page 451 of \cite{PutermanMDP}, we know that there exists an optimal policy that is deterministic. Therefore, we only need to show that for any arbitrary deterministic stationary policy $\pi$, its long-run average daily reward $R(\pi \vert s)$ is upper bounded by the fluid-based LP optimal objective value $\bar{R}$. Denote $(\jointtripfulfillaction{}{}(s^t), \jointreroutingaction{}{}(s^t), \jointchargingaction{}{}(s^t), \jointidlingaction{}{}(s^t), \jointpassaction{}{}(s^t))$ as a  deterministic flow of vehicles generated by policy $\pi$ given state $s^t$ at time step $t$. Since the state space of the MDP is finite, we know that given any initial state $s \in \mathcal{S}$ and any deterministic stationary policy $\pi$, the system has a stationary state distribution, denoted as $\rho$. Here, the dependence of $\rho$ on the policy $\pi$ and the initial state $s$ is dropped for notation simplicity. By proposition 8.1.1 in page 333 of \cite{PutermanMDP}, the long-run average daily reward defined in \eqref{eq:long-run-avg-reward} can be written as follows: 
    \begin{align*}
        &R(\pi|s) = \mathbb{E}_{\rho}  \left[\sum_{\time \in [\Horizon]}^{} \sum_{\cartype{} \in \Cartype} \sum_{(\origin, \destination) \in \Region \times \Region} \right. \\
        &\qquad\  \left(\tripfulfillreward{,\origin\destination}{\time}  \sum_{\tripactivetime \in [\Lc]}^{} \jointtripfulfillaction{\cartype{}, (\origin, \destination, \xi)}{}(s^{\time}) + \reroutingreward{,\origin\destination}{\time} \jointreroutingaction{\cartype{}, \destination}{}(s^t) \right)\\ 
        &\quad+ \left. \sum_{\cartype{} \in \Cartype}\sum_{\type \in \Type} \chargingreward{,\type}{\time} \jointchargingaction{\cartype{},\type}{}(s^t) \right].
    \end{align*}
    Given $(\jointtripfulfillaction{}{}(s^t), \jointreroutingaction{}{}(s^t), \jointchargingaction{}{}(s^t), \jointidlingaction{}{}(s^t), \jointpassaction{}{}(s^t))$ for each state $s^t$ at each time step $t \in [\Horizon]$, we define the fleet dispatch fraction vector $(\lptripfulfill{}{t}, \lpreroute{}{t}, \lpcharge{}{t}, \lppass{}{t})$ for each $t \in [T]$:
    \begin{enumerate}
        \item[(i)] $\lptripfulfill{\cartype{}, \triptype{}}{t} := \frac{1}{\Size} \mathbb{E}_{\rho}[ \jointtripfulfillaction{\cartype{}, \triptype{}}{}(s^t)]$, $\forall \cartype{} \in \Cartype$, $\triptype{} \in \Triptype$, and $\time \in [\Horizon]$,
        \item[(ii)] $\lpreroute{\cartype{}, \region}{t} := \frac{1}{\Size} \mathbb{E}_{\rho}[\jointreroutingaction{(\origin, 0, \battery), \region}{}(s^t)]$, $\forall \cartype{} := (\origin, 0, \battery) \in \Cartype$, $\region \in \Region$, $\region \neq \origin$, and $\time \in [\Horizon]$,
        \item[(iii)] $\lpreroute{(\region, 0, \battery), \region}{t} := \frac{1}{\Size} \mathbb{E}_{\rho}[\jointpassaction{(\region, 0, \battery)}{}(s^t)]$, $\forall \cartype{} := (\region, 0, \battery) \in \Cartype$, $\region \in \Region$, and $\time \in [\Horizon]$,
        \item[(iv)] $\lpcharge{\cartype{}, \type}{t} := \frac{1}{\Size} \mathbb{E}_{\rho}[\jointchargingaction{\cartype{}, \type}{}(s^t)]$, $\forall \cartype{} \in \Cartype$, $\type \in \Type$, and $\time \in [\Horizon]$,
        \item[(v)] $\lppass{\cartype{}}{t} := \frac{1}{\Size} \mathbb{E}_{\rho}[\jointpassaction{\cartype{}}{}(s^t)]$, $\forall \cartype{} \in \Cartype$ and $\time \in [\Horizon]$,
    \end{enumerate}
    where the expectations are taken over the states $s^t$. First, given that $(\jointtripfulfillaction{}{}(s^t), \jointreroutingaction{}{}(s^t), \jointchargingaction{}{}(s^t), \jointidlingaction{}{}(s^t), \jointpassaction{}{}(s^t))$ is a feasible fleet flow that satisfies all constraints in Sec. \ref{sec:model}, we show that $(\lptripfulfill{}{}, \lpreroute{}{}, \lpcharge{}{}, \lppass{}{})$ is a feasible solution to the fluid-based LP.

    We show that constraint \eqref{eq:fluid-ev-conservation-red} holds. The left-hand side aggregates all vehicles from $t-1$ that transitions to status $\cartype{}$, while the right-hand side aggregates vehicles of status $\cartype{}$ assigned to take all actions at time $t$. Both sides equal to the fraction of vehicles with status $c$ at time $t$, and thus \eqref{eq:fluid-ev-conservation-red} holds.

    Next, we show that constraint \eqref{eq:fluid-passenger-flow-cap-red} holds. By \eqref{eq:setup-trip-state-transition}, we obtain that for any $\origin, \destination \in \Region$ and any $t \in [\Horizon]$, 
    \begin{align*}
        0 \stackrel{(a)}{\leq}& \state{(\origin, \destination, \Lc)}{\time} - \sum_{\cartype{} \in \Cartype} \jointtripfulfillaction{\cartype{}, (\origin, \destination, \Lc)}{}(s^t)\\
        \stackrel{(b)}{=}& \state{(\origin, \destination, \Lc - 1)}{\time - 1} - \sum_{\cartype{} \in \Cartype} \left(\jointtripfulfillaction{\cartype{}, (\origin, \destination, \Lc)}{}(s^t) + \jointtripfulfillaction{\cartype{}, (\origin, \destination, \Lc - 1)}{}(s^{t-1}) \right)\\
        =& \dots\\
        \stackrel{(c)}{=}& \state{(\origin, \destination, 0)}{\time - \Lc} - \sum_{\cartype{} \in \Cartype} \sum_{\tripactivetime \in [\Lc]} \jointtripfulfillaction{\cartype{}, (\origin, \destination, \Lc - \tripactivetime)}{}(s^{\time - \tripactivetime})\\
        \stackrel{(d)}{\leq}& \arrnum{\origin\destination}{\time - \Lc} - \sum_{\cartype{} \in \Cartype} \sum_{\tripactivetime \in [\Lc]} \jointtripfulfillaction{\cartype{}, (\origin, \destination, \Lc - \tripactivetime)}{}(s^{\time - \tripactivetime}),
    \end{align*}
    where (a) is due to constraint \eqref{eq:setup-trip-fulfill-cap}; (b) and (c) are obtained by \eqref{eq:setup-trip-state-transition} under the condition that $1 \leq \tripactivetime \leq \Lc$; and (d) is because the trip state $\state{(\origin, \destination, 0)}{\time - \Lc}$ is upper bounded by $\arrnum{\origin\destination}{\time - \Lc}$ as in \eqref{eq:setup-trip-state-transition} for $\tripactivetime = 0$.
    Thus, we obtain 
    \begin{align*}
        \arrnum{\origin\destination}{\time - \Lc} \geq& \sum_{\cartype{} \in \Cartype} \sum_{\tripactivetime \in [\Lc]} \jointtripfulfillaction{\cartype{}, (\origin, \destination, \Lc - \tripactivetime)}{}(s^{\time - \tripactivetime}) \\
        =& \sum_{\battery \in [\range]} \sum_{\timetoarrival \in [\Lp]} \sum_{\tripactivetime \in [\Lc]} \jointtripfulfillaction{(\origin, \timetoarrival, \battery), (\origin, \destination, \Lc - \tripactivetime)}{}(s^{\time - \tripactivetime}).
    \end{align*}
We shift $t$ to be $t+L_c$, then the above inequality can be re-written as follows: \begin{align*}
        &\frac{1}{\Size} \arrnum{\origin\destination}{\time} \\\geq& \frac{1}{\Size} \sum_{\battery \in [\range]} \sum_{\timetoarrival \in [\Lp]} \sum_{\tripactivetime \in [\Lc]} \jointtripfulfillaction{(\origin, \timetoarrival, \battery), (\origin, \destination, L_c- \tripactivetime)}{}(s^{\time + L_c - \tripactivetime})\\
        =&\frac{1}{\Size} \sum_{\battery \in [\range]} \sum_{\timetoarrival \in [\Lp]} \sum_{\tripactivetime \in [\Lc]} \jointtripfulfillaction{(\origin, \timetoarrival, \battery), (\origin, \destination, \tripactivetime)}{}(s^{\time + \tripactivetime}), 
    \end{align*}
    where the second equation is due to swapping $L_c - \xi$ as $\xi$. Taking expectation on both sides, we obtain
    \begin{align*}
        \frac{1}{\Size} \arrrate{\origin \destination}{\time} =& \mathbb{E}_{\rho}\left[\frac{1}{\Size} \arrnum{\origin\destination}{\time}(s^t) \right]\\
        \geq& \frac{1}{\Size} \mathbb{E}_{\rho}\left[\sum_{\battery \in [\range]} \sum_{\timetoarrival \in [\Lp]} \sum_{\tripactivetime \in [\Lc]} \jointtripfulfillaction{(\origin, \timetoarrival, \battery), (\origin, \destination, \tripactivetime)}{}(s^{\time + \tripactivetime}) \right]\\
        =& \sum_{\battery \in [\range]} \sum_{\timetoarrival \in [\Lp]} \sum_{\tripactivetime \in [\Lc]} \mathbb{E}_{\rho}\left[\frac{1}{\Size} \jointtripfulfillaction{(\origin, \timetoarrival, \battery), (\origin, \destination, \tripactivetime)}{}(s^{\time + \tripactivetime})\right]\\
        =& \sum_{\cartype{} = (\origin, \timetoarrival, \battery) \in \Cartype} \sum_{\triptype{} = (\origin, \destination, \tripactivetime) \in \Triptype} \lptripfulfill{\cartype{}, \triptype{}}{\time + \tripactivetime}.
    \end{align*}

    The constraint \eqref{eq:fluid-charging-cap-red} holds because:
    \begin{align*}
        &\sum_{\plugremaintime \in [\chargetime]} \sum_{\cartype{} = (\region, 0, \battery) \in \Cartype} \lpcharge{\cartype{}, \type}{t-\plugremaintime} \\
        =& \frac{1}{\Size} \mathbb{E}_{\rho} \left[\sum_{\plugremaintime \in [\chargetime]} \sum_{\cartype{} = (\region, 0, \battery) \in \Cartype} \jointchargingaction{\cartype{}, \type}{}(s^{t-\plugremaintime})\right] \\
        \stackrel{(a)}{=}& \frac{1}{\Size} \sum_{\plugremaintime = 0}^{\chargetime - 1} s_{(\origin, \type, \chargetime - 1)}^{t - \plugremaintime}\\
        \stackrel{(b)}{=}& \frac{1}{\Size} s_{(\origin, \type, 1)}^{t-1} + \frac{1}{\Size} \sum_{\plugremaintime = 1}^{\chargetime - 1} s_{(\origin, \type, \plugremaintime)}^{t}\\
        \stackrel{(c)}{\leq}& \frac{1}{\Size} s_{(\origin, \type, 0)}^{t} + \frac{1}{\Size} \sum_{\plugremaintime = 1}^{\chargetime - 1} s_{(\origin, \type, \plugremaintime)}^{t}\\
        \stackrel{(d)}{=}& \frac{1}{\Size} \n{\region}{\type},
    \end{align*}
    where (a) is obtained by the case of $\plugremaintime = \chargetime - 1$ in \eqref{eq:setup-plug-state-transition}; (b) is obtained by the case of $0 < \plugremaintime < \chargetime - 1$ in \eqref{eq:setup-plug-state-transition}; (c) is obtained by the case of $\plugremaintime = 0$ in \eqref{eq:setup-plug-state-transition} and the constraint in \eqref{eq:setup-charging-cap}; and (d) is obtained by \eqref{eq:setup-charger-total-num}.

    Constraints \eqref{eq:fluid-passenger-flow-battery-sufficiency-red} and \eqref{eq:fluid-rerouting-flow-battery-sufficiency-red} can be obtained by the constraints \eqref{eq:setup-passenger-carrying-flow} and \eqref{eq:setup-rerouting-flow} in Sec. \ref{sec:model}. Constraint \eqref{eq:fluid-total-flow-red} holds due to \eqref{eq:setup-car-total-flow}. Finally, constraint \eqref{eq:fluid-nonneg-red} holds because the number of fleet assigned to each action must be non-negative. Hence, we obtain that $(\lptripfulfill{}{}(s^t), \lpreroute{}{}(s^t), \lpcharge{}{}(s^t), \lppass{}{}(s^t))_{s^t \in S, t \in [T]}$ is a feasible solution of the fluid-based LP \eqref{eq:fluid_LP}.
    
    Therefore, we have
    \begin{align*}
        &R(\pi \vert s) = \mathbb{E}_{\rho}  \left[\sum_{\time \in [\Horizon]}^{} \sum_{\cartype{} \in \Cartype} \sum_{(\origin, \destination) \in \Region \times \Region} \right. \\
        &\qquad\  \left(\tripfulfillreward{,\origin\destination}{\time}  \sum_{\tripactivetime \in [\Lc]}^{} \jointtripfulfillaction{\cartype{}, (\origin, \destination, \xi)}{}(s^{\time}) + \reroutingreward{,\origin\destination}{\time} \jointreroutingaction{\cartype{}, \destination}{}(s^t) \right)\\ 
        &\quad+ \left. \sum_{\cartype{} \in \Cartype}\sum_{\type \in \Type} \chargingreward{,\type}{\time} \jointchargingaction{\cartype{},\type}{}(s^t) \right]\\
        \stackrel{(a)}{=}& \sum_{\time \in [\Horizon]}^{} \sum_{\cartype{} \in \Cartype} \sum_{(\origin, \destination) \in \Region \times \Region} \\
        &\qquad\ \left( \tripfulfillreward{,\origin\destination}{\time} \sum_{\tripactivetime \in [\Lc]}^{} \mathbb{E}_{\rho}  \left[\jointtripfulfillaction{\cartype{}, (\origin, \destination, \xi)}{}(s^{\time}) \right] \right. \\ 
        &\quad+ \left. \reroutingreward{,\origin\destination}{\time} \mathbb{E}_{\rho}  \left[\jointreroutingaction{\cartype{}, \destination}{}(s^t) \right]\right) \\ 
        &\quad+ \sum_{\cartype{} \in \Cartype}\sum_{\type \in \Type} \chargingreward{,\type}{\time} \mathbb{E}_{\rho}  \left[\jointchargingaction{\cartype{},\type}{}(s^t) \right]\\
        \stackrel{(b)}{=}& \Size \cdot \left[\sum_{\time \in [\Horizon]}^{} \sum_{\cartype{} \in \Cartype} \sum_{(\origin, \destination) \in \Region \times \Region} \right. \\
        &\qquad\  \left(\tripfulfillreward{,\origin\destination}{\time}  \sum_{\tripactivetime \in [\Lc]}^{} \lptripfulfill{\cartype{}, (\origin, \destination, \xi)}{\time} + \reroutingreward{,\origin\destination}{\time} \lpreroute{\cartype{}, \destination}{\time} \right)\\ 
        &\quad+ \left. \sum_{\cartype{} \in \Cartype}\sum_{\type \in \Type} \chargingreward{,\type}{\time} \lpcharge{\cartype{},\type}{\time} \right]\\
        \stackrel{(c)}{\leq}& \bar{R}, 
    \end{align*}
    where (a) is obtained by the linearity of expectation, (b) is by plugging in (i)-(v) above, (c) is because $(\lptripfulfill{}{}, \lpreroute{}{}, \lpslack{}{}, \lpcharge{}{}, \lppass{}{})$ is a feasible solution to the fluid-based LP. As a result, the associated reward function value is upper bounded by the optimal value $\bar{R}$.
    \hfill $\square$

\subsection{Reduction of Fluid Based Linear Program}\label{apx:reduction}
The original fluid-based linear program has $\vert \Region \vert (\Lp + \maxtimecost{}{}) \range \Horizon$ number of variables, where $\maxtimecost{}{}$ can be large. In this section, we reformulate \eqref{eq:fluid_LP} to obtain a linear program with fewer decision variables and constraints, reducing the number of variables to $|V|L_pBT$ without loss of optimality. This reformulation is achieved based on the fact that only vehicles with task remaining time $\timetoarrival \leq \Lp$ can be assigned with new tasks, and therefore we only need to keep track of the statuses of vehicles with task remaining time $\timetoarrival \leq \Lp$ in the optimization formulation. We define the variables of the simplified fluid-based optimization problem as follows:
\begin{enumerate}
    \item[-] \emph{Fraction of fleet for trip fulfilling $\lpredtripfulfill{}{} := \left(\lpredtripfulfill{\origin \destination, \battery, \timetoarrival}{\time}\right)_{\origin,\destination \in \Region, \time \in [T], \battery \in [B], \timetoarrival \in[\Lp]}$}, where $\lpredtripfulfill{\origin \destination, \battery, \eta}{\time}$ is the fraction of vehicles with battery level $\battery$ at time $\time$ that pick up trip requests from $\origin$ to $\destination$ with remaining waiting time $\timetoarrival$. We remark two differences between $\lpredtripfulfill{}{}$ and  $\lptripfulfill{}{}$ in the original formulation: (i) $\lpredtripfulfill{}{}$ only keeps track of vehicles with task remaining time $\timetoarrival \leq \Lp$, whereas $\lptripfulfill{}{}$ keeps track of all trip-fulfilling vehicles; (ii) $\lpredtripfulfill{}{}$ aggregates $\lptripfulfill{}{}$ across all trip active times $\tripactivetime$, i.e., $\lpredtripfulfill{\origin \destination, \battery, \timetoarrival}{\time} = \sum_{\tripactivetime \in [\Lc]} \lptripfulfill{(\origin, \timetoarrival, \battery), (\origin, \destination, \tripactivetime)}{\time}$.
    \item[-] \emph{Fraction of fleet fulfilling trip requests with a certain status $\lpredslack{}{} := \left(\lpredslack{\origin \destination, \timetoarrival, \tripactivetime}{\time}\right)_{\origin,\destination \in \Region, \eta \in [\Lp], \tripactivetime \in [\Lc], \time \in [T]}$}, where $\lpredslack{\origin \destination, \timetoarrival, \tripactivetime}{\time}$ is fraction of vehicles assigned at time $t$ to a trip with o-d pair $\origin\destination$ that has been waiting for assignment for $\tripactivetime$ steps and will be picked up after $\timetoarrival$ time steps. We note that $\lpredslack{}{}$ differs from $\lpredtripfulfill{}{}$ in that $\lpredslack{}{}$ aggregates the trip-fulfilling vehicles across battery level $\battery$, while $\lpredtripfulfill{}{}$ aggregates them across trip active time $\tripactivetime$. In particular, $\lpslack{\origin \destination, \timetoarrival, \tripactivetime}{\time} = \sum_{\battery \in [\range]} \lptripfulfill{(\origin, \timetoarrival, \battery), (\origin, \destination, \tripactivetime)}{\time}$.
    \item[-] \emph{Fraction of repositioning fleet $\lpredreroute{}{} := \left(\lpredreroute{\origin \destination, \battery}{\time}\right)_{\origin,\destination \in \Region, \time \in [T], \battery \in [B]}$}, where $\lpredreroute{\origin \destination, \battery}{\time}$ is the fraction of vehicles with battery level $\battery$ at time $\time$ repositioning from $\origin$ to $\destination$. Here, the variable $\lpredreroute{\origin \destination, \battery}{\time}$ is equivalent to the decision variable $\lpreroute{(\origin, 0, \battery), \destination}{\time}$ in the original formulation.
    \item[-] \emph{Fraction of charging fleet $\lpredcharge{}{} := \left(\lpredcharge{\region, \type,\battery}{\time}\right)_{\region \in \Region, \type \in \Type, \time \in [T], \battery \in [\range]}$}, where $\lpredcharge{\region, \type, \battery}{\time}$ denotes the fraction of vehicles with battery level $\battery$ charging at region $\region$ with rate $\type$ at time $\time$. Here, the variable $\lpredcharge{\region, \type, \battery}{\time}$ is equivalent to the decision variable $\lpcharge{(\region, 0, \battery), \type}{\time}$ in the original formulation.
    \item[-] \emph{Fraction of fleet for passing action $\lpredpass{}{} := \left(\lpredpass{\region, \battery, \timetoarrival}{\time}\right)_{\region \in \Region, \battery \in[\range], \timetoarrival \in [\Lp], \time \in [\Horizon]}$}, where $\lpredpass{\region, \battery, \timetoarrival}{\time}$ is the fraction of fleet $\timetoarrival$ time steps away from destination $\region$ with battery $\battery$ that takes the passing action at time $\time$. Here, the variable $\lpredpass{\region, \battery, \timetoarrival}{\time}$ is equivalent to the decision variable $\lppass{(\region, \timetoarrival, \battery)}{\time}$ in the original formulation. Again, we note that $\lpredpass{}{}$ only keeps track of the vehicles statuses with task remaining time $\timetoarrival \leq \Lp$, while $\lppass{}{}$ in the original formulation keeps track of all vehicle statuses.
\end{enumerate}

The simplified fluid-based linear program is given as follows: 

\allowdisplaybreaks
{\small \begin{subequations} \label{eq:fluid-lp}
\begin{align}
    &\max_{\lpredtripfulfill{}{}, \lpredreroute{}{}, \lpredcharge{}{}, \lpredslack{}{}, \lpredpass{}{}} \  \  \Size \sum_{\time \in [\Horizon]}^{} \left\{ \sum_{\origin \in \Region} \sum_{\destination \in \Region} \left[ \tripfulfillreward{,\origin\destination}{\time} \right. \right. \nonumber \\
    &\qquad \left. \left(\sum_{\tripactivetime \in [\Lc]}^{}\sum_{\timetoarrival \in [\Lp]}^{} \lpredslack{\origin \destination, \timetoarrival, \tripactivetime}{\time} \right) + \reroutingreward{,\origin\destination}{\time} \sum_{\battery \in [\range]}^{} \lpredreroute{\origin \destination, \battery}{\time} \right] \nonumber\\ 
    &\qquad+ \left. \sum_{\type \in \Type} \chargingreward{,\type}{\time} \sum_{\region \in \Region} \sum_{\battery \in [\range]}^{} \lpredcharge{\region, \type, \battery}{\time} \right\}, \notag \\
    &\text{s.t.} \  \left[ \sum_{\destination \in \Region} \left( \sum_{\timetoarrival' \in [\Lp]}^{} \lpredtripfulfill{\destination \origin, \battery + \batterycost{\destination \origin}, \timetoarrival'}{\phi_{\destination \origin}(t, \timetoarrival')} \nonumber \right. \right.\\ 
    &\qquad+ \left. \left. \lpredreroute{\destination \origin, \battery + \batterycost{\destination \origin}}{\phi_{\destination \origin}(t, 0)} \mathds{1}\{\destination \neq \origin\} \right) \mathds{1}\{\battery + \batterycost{\destination \origin} \leq \range \} \right. \nonumber \\
    &\qquad+ \left. \sum_{\type \in \Type} \lpredcharge{\origin, \type, \battery - \type \chargetime + \batterycost{\origin \origin}}{\phi_{\origin \origin}^{\time - \chargetime}} \mathds{1}\{\battery \geq \type \chargetime - \batterycost{\origin \origin}\} \right. \notag\\ 
    &\qquad+ \left. \sum_{\type \in \Type} \sum_{\battery' > \range - \type \chargetime + \batterycost{\origin \origin}} \lpredcharge{\origin, \type, \battery'}{\phi_{\origin \origin}^{\time - \chargetime}} \mathds{1}\{\battery = \range\} \right] \mathds{1}\{\timetoarrival = \Lp\} \nonumber\\ 
    &\qquad+ \lpredreroute{\origin \origin, \battery}{(\time - 1)} \mathds{1}\{\timetoarrival = 0\} + \lpredpass{\origin, \battery, \timetoarrival + 1}{(\time - 1)} \mathds{1}\{\timetoarrival < \Lp\} \notag \\
    &\quad= \sum_{\destination \in \Region} \left(\lpredtripfulfill{\origin \destination, \battery, \timetoarrival}{\time} + \lpredreroute{\origin \destination, \battery}{\time} \right) + \sum_{\type \in \Type} \lpredcharge{\origin, \type, \battery}{\time} + \lpredpass{\origin, \battery, \timetoarrival}{\time} \mathds{1}\{\timetoarrival > 0\},\nonumber\\ 
    &\qquad \forall \origin \in \Region, \   \timetoarrival \in [\Lp], \   \time \in [\Horizon], \  \battery \in [\range], \label{eq:fluid-ev-conservation}
\end{align}
\begin{align}
    &\ \sum_{\battery \in [\range]}^{} \lpredtripfulfill{\origin \destination, \battery, \timetoarrival}{\time} = \sum_{\tripactivetime \in [\Lc]}^{} \lpredslack{\origin \destination, \timetoarrival, \tripactivetime}{\time}, \nonumber\\
    &\qquad \forall \origin, \destination \in \Region, \  \timetoarrival \in [\Lp], \  \time \in [\Horizon], \label{eq:fluid-passenger-flow-decomp}\\
    &\ \sum_{\tripactivetime \in [\Lc]}^{}\sum_{\timetoarrival \in [\Lp]}^{} \lpredslack{\origin \destination, \timetoarrival, \tripactivetime}{(\time + \tripactivetime)} \leq \frac{1}{\Size} \arrrate{\origin \destination}{\time}, \quad \forall \origin, \destination \in \Region, \ \time \in [\Horizon],\label{eq:fluid-passenger-flow-cap}\\
    &\ \sum_{j \in [\chargetime]} \sum_{\battery \in [\range]}^{} \lpredcharge{\region, \type, \battery}{(\time - j)} \leq \frac{1}{\Size} \n{\region}{\type}, \quad \forall \type \in \Type, \  \region \in \Region, \  \time \in [\Horizon], \label{eq:fluid-charging-cap}\\
    &\ \lpredtripfulfill{\origin \destination, \battery, \timetoarrival}{\time} \mathds{1}\{\battery < \batterycost{\origin \destination}\} = 0,\notag\\ 
    &\quad \forall \battery \in [\range], \  \origin, \destination \in \Region, \  \time \in [\Horizon], \  \timetoarrival \in [\Lp], \label{eq:fluid-passenger-flow-battery-sufficiency}\\
    &\ \lpredreroute{\origin \destination, \battery}{\time} \mathds{1}\{\battery < \batterycost{\origin \destination}\} = 0,\notag\\ 
    &\quad \forall \battery \in [\range], \  \origin, \destination \in \Region, \  \time \in [\Horizon], \label{eq:fluid-rerouting-flow-battery-sufficiency} \\
    &\ \sum_{\battery \in [\range]} \left[ \sum_{\origin, \destination \in \Region} \left(\sum_{\timetoarrival' \in [\Lp]}^{} \sum_{t' \in \psi_{\origin \destination}(t, \eta')}\lpredtripfulfill{\origin \destination, b, \eta'}{t'}  \right. \right. + \nonumber\\ 
    &\qquad \left. \sum_{t' \in \psi_{\origin \destination}(t, 0)} \lpredreroute{\origin \destination, b}{t'}\right) \nonumber\\ 
    &\quad+ \left. \sum_{\region \in \Region} \sum_{\type \in \Type} \sum_{t' = t-\chargetime + \Lp}^t \lpredcharge{\region,\type,\battery}{t'} + \sum_{\region \in \Region} \sum_{\timetoarrival \in [\Lp]}^{} \lpredpass{\region, \battery, \timetoarrival}{\time} \right] = 1, \nonumber\\
    &\qquad \forall \time \in [\Horizon], \label{eq:fluid-total-flow}\\
    &\ \lpredtripfulfill{}{}, \lpredreroute{}{}, \lpredcharge{}{}, \lpredslack{}{}, \lpredpass{}{} \geq 0, \label{eq:fluid-nonneg}
\end{align}    
\end{subequations}
}where $\phi_{\origin \destination}(t, \timetoarrival')$ is defined as the time at which the vehicle assigned to a trip from region $\origin$ to region $\destination$ will be $\Lp$ time steps away from $v$ at time $t$:
\begin{align}\label{eq:fluid-phi}
    &\phi_{uv}(t, \eta') + \eta' + \tau_{uv}^{\phi_{uv}(t, \eta') + \eta'} - \Lp = t, \nonumber\\
    &\quad \forall u, v \in \Region,\  \forall \eta' \in [\Lp], \  \forall t \in [T],
\end{align}
and $\psi_{\origin \destination}(t, \eta')$ is defined as the set of times at which the vehicle assigned to a trip from $\origin$ to $\destination$ will be {\em at least} $\Lp$ time steps away from $v$ at time $t$:
\begin{align} \label{eq:fluid-psi}
    \psi_{\origin \destination}(t, \eta') =& \left\{t' \in [T] \lvert t' \leq t, \right. \nonumber \\ 
    &\left. t'+ \eta' + \tau_{uv}^{t'+\eta'} - \Lp \geq t \right\}.
\end{align}

The objective function represents the daily average reward achieved by the vehicle flow. We argue that this objective function is equivalent to the objective function of the original formulation. For each pair of origin $\origin$ and destination $\destination$ at time $\time$, the term $\sum_{\tripactivetime \in [\Lc]} \sum_{\timetoarrival \in [\Lp]} \lpredslack{\origin \destination, \timetoarrival, \tripactivetime}{\time}$ aggregates all vehicles that are assigned to fulfill trip requests from $\origin$ to $\destination$ at time $\time$, which is equivalent to the term $\sum_{\cartype{} \in \Cartype} \sum_{\tripactivetime \in [\Lc]} \lptripfulfill{\cartype{}, (\origin, \destination, \tripactivetime)}{\time}$ in the objective function of the original formulation. The term $\sum_{\battery \in [\range]} \lpredreroute{\origin \destination, \battery}{\time}$ aggregates all vehicles assigned to reposition from $\origin$ to $\destination$ at time $\time$, which is equivalent to the term $\sum_{\cartype{} \in \Cartype} \lpreroute{\cartype{}, \destination}{\time}$ in the objective function of the original formulation. Finally, the term $\sum_{\battery \in [\range]} \lpredcharge{\region, \type, \battery}{\time}$ aggregates all vehicles assigned to charge at region $\region$ with rate $\type$ at time $\time$, which is equivalent to the term $\sum_{\cartype{} \in \Cartype} \lpcharge{\cartype{}, \type}{\time}$ in the objective function of the original formulation.

Constraint \eqref{eq:fluid-ev-conservation} ensures the flow conservation for each vehicle status $(\origin, \timetoarrival, \battery)$ with $\timetoarrival \leq \Lp$ at any time $\time$ of day. We remark that the constraint \eqref{eq:fluid-ev-conservation-red} in the original formulation also includes the flow conservation for vehicles statuses with $\timetoarrival > \Lp$. We can reduce these constraints because passing is the only feasible actions for these vehicles. 
The left-hand side of the equation represents the vehicles being assigned at previous times that reach the status $(\origin, \timetoarrival, \battery)$ at time $\time$, which is equivalent to the left-hand side of \eqref{eq:fluid-ev-conservation-red} in the original formulation and can be obtained by the vehicle state transition equation \eqref{eq:setup-car-state-transition}. In particular, the left-hand side includes the following terms: (i) the term $\lpredtripfulfill{\destination \origin, \battery + \batterycost{\destination \origin}, \timetoarrival'}{\phi_{\destination \origin}(t, \timetoarrival')}$ represents that vehicles of status $(\destination, \battery + \batterycost{\destination \origin}, \timetoarrival')$ at time $\phi_{\destination \origin}(\time, \timetoarrival')$ become vehicles of status $(\origin, \battery, \timetoarrival)$ after being assigned to pick up trip orders from $\destination$ to $\origin$; (ii) the term $\lpredreroute{\destination \origin, \battery + \batterycost{\destination \origin}}{\phi_{\destination \origin}(t, 0)}$ represents that vehicles of status $(\destination, \battery + \batterycost{\destination \origin}, 0)$ at time $\phi_{\destination \origin}(\time, 0)$ become vehicles of status $(\origin, \battery, \timetoarrival)$ after being assigned to reposition from $\destination$ to $\origin$ if $\destination \neq \origin$. On the other hand, when $\timetoarrival = 0$, the term $\lpredreroute{\origin \origin, \battery}{(\time - 1)} \mathds{1}\{\timetoarrival = 0\}$ represents that vehicles of status $(\origin, 0, \battery)$ at time $\time - 1$ remain the same status at time $\time$ after being assigned to idle; (iii) the term $\lpredcharge{\origin, \type, \battery - \type \chargetime + \batterycost{\origin \origin}}{\phi_{\origin \origin}^{\time - \chargetime}}$ represents that vehicles of status $(\destination, \battery - \batterycost{\destination \origin} + \batterycost{\origin \origin}, 0)$ at time $\phi_{\origin \origin}^{\time - \chargetime}$ become vehicles of status $(\origin, \battery, \timetoarrival)$ after being assigned to charge in region $\origin$ at rate $\type$. Moreover, the term $\sum_{\battery' > \range - \type \chargetime + \batterycost{\origin \origin}} \lpredcharge{\origin, \type, \battery'}{\phi_{\origin \origin}^{\time - \chargetime}}$ represents that if $\battery = \range$, all vehicles of status $(\origin, \battery', 0)$ with battery levels $\battery' > \range - \type \chargetime + \batterycost{\origin \origin}$ will charge to full and become vehicles of status $(\origin, \range, \timetoarrival)$ after being assigned to charge in region $\origin$ at rate $\type$; (iv) the term $\lpredpass{\origin, \battery, \timetoarrival + 1}{(\time - 1)}$ represents that vehicles of status $(\region, \battery, \timetoarrival + 1)$ assigned a passing action at time $\time - 1$ become vehicles of status $(\region, \battery, \timetoarrival)$ at time $\time$. The right hand side of \eqref{eq:fluid-ev-conservation} represents the outgoing vehicle flows for each vehicle status, which can be obtained by \eqref{eq:setup-cartype-conservation} and is equivalent to the right-hand side of \eqref{eq:fluid-ev-conservation-red}. 

Constraint \eqref{eq:fluid-passenger-flow-decomp} represents the equivalence of two aggregations of trip-fulfilling vehicle flows, which is reflected by the definition of $\lpredtripfulfill{}{}$ and $\lpredslack{}{}$ at the beginning of this section. 
Constraint \eqref{eq:fluid-passenger-flow-cap} ensures that the fulfillment of trip orders does not exceed their arrivals, which is equivalent to the constraint \eqref{eq:fluid-passenger-flow-cap-red} in the original formulation. 
Constraints \eqref{eq:fluid-passenger-flow-battery-sufficiency} and \eqref{eq:fluid-rerouting-flow-battery-sufficiency} ensure that vehicles will not travel to some regions if their battery levels are below the battery cost required to complete the trip, which are equivalent to the constraints \eqref{eq:fluid-passenger-flow-battery-sufficiency-red} and \eqref{eq:fluid-rerouting-flow-battery-sufficiency-red} in the original formulation. 

Constraint \eqref{eq:fluid-total-flow} ensures that the total fraction of vehicles should add up to $1$ at all times and is equivalent to the constraint \eqref{eq:fluid-total-flow-red} in the original formulation. At each time, a vehicle can either be dispatched to pick up trip requests, reposition, charge, pass, continue with ongoing trips, or continue being charged. $\psi_{\origin \destination}(t, \timetoarrival')$ represents the time steps at which vehicles $\timetoarrival'$ away from $\origin$ and assigned to pick up trip requests from $\origin$ to $\destination$ will still be more than $\Lp$ time steps away from reaching $\destination$ at time $\time$. In other words, these vehicles are not feasible for non-passing actions at time $\time$. In particular, the constraint \eqref{eq:fluid-total-flow} includes the following terms: (i) the term $\sum_{\timetoarrival' \in [\Lp]}^{} \sum_{t' \in \psi_{\origin \destination}(t, \eta')}\lpredtripfulfill{\origin \destination, b, \eta'}{t'}$ represents all vehicles assigned to pickup trip requests from $\origin$ to $\destination$ that are still more than $\Lp$ time steps away from reaching $\destination$ at time $\time$; (ii) the term $\sum_{t' \in \psi_{\origin \destination}(t, 0)} \lpredreroute{\origin \destination, b}{t'}$ represents all vehicles assigned to reposition from $\origin$ to $\destination$ that are still more than $\Lp$ time steps away from reaching $\destination$ at time $\time$; (iii) the term $\sum_{t' = t-\chargetime + \Lp}^t \lpredcharge{\region,\type,\battery}{t'}$ represents all vehicles assigned to charge in region $\region$ with rate $\type$ that are still more than $\Lp$ time steps away from completing their charging periods; (iv) the term $\sum_{\region \in \Region} \sum_{\timetoarrival \in [\Lp]}^{} \lpredpass{\region, \battery, \timetoarrival}{\time}$ represents all vehicles assigned to pass at time $\time$. 

The constraint \eqref{eq:fluid-nonneg} requires that all decision variables should be non-negative, which is equivalent to the constraint \eqref{eq:fluid-nonneg-red} in the original formulation.

\clearpage




%
%
%


\bibliographystyle{unsrtnat} 
\bibliography{references} 

\end{document}